\documentclass[11pt]{article}

\usepackage[preprint]{coling}

\usepackage{times}
\usepackage{latexsym}

\usepackage[T1]{fontenc}

\usepackage[utf8]{inputenc}

\usepackage{microtype}

\usepackage{inconsolata}
\usepackage{coling}
\usepackage{graphicx}

\usepackage{amssymb}
\usepackage{amsfonts}
\usepackage{bm}       
\usepackage{amsmath}  
\usepackage{enumitem}  
\usepackage{pifont}

\usepackage{enumitem}

\usepackage{booktabs} 
\usepackage{color}
\usepackage{float}
\usepackage{CJKutf8}
\usepackage{booktabs} 
\usepackage{multirow}
\usepackage{tabularx}
\usepackage{makecell}
\usepackage{siunitx}

\def\mylogo{\resizebox{0.45cm}{!}{\includegraphics{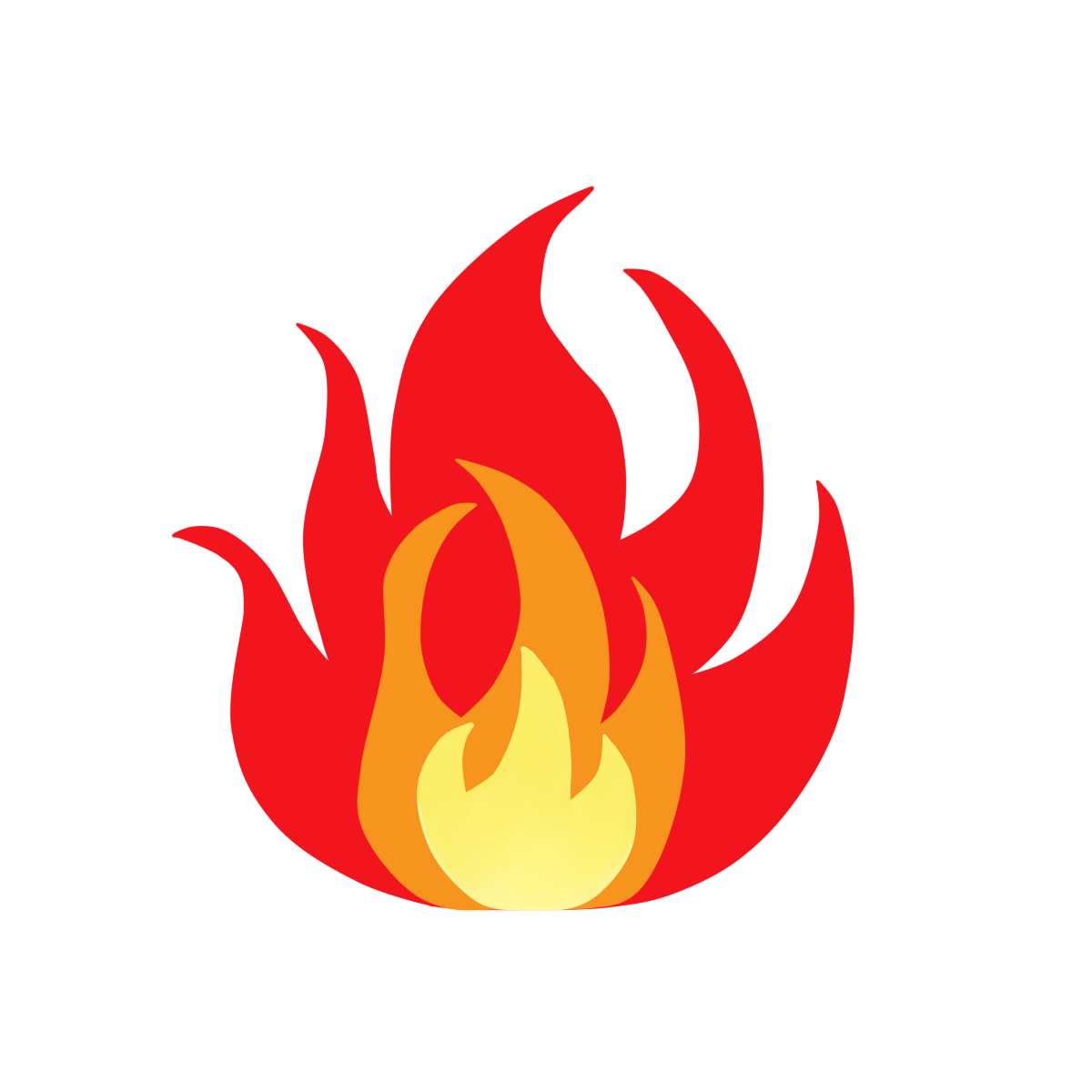}}} 
\def\mylogosnow{\resizebox{0.45cm}{!}{\includegraphics{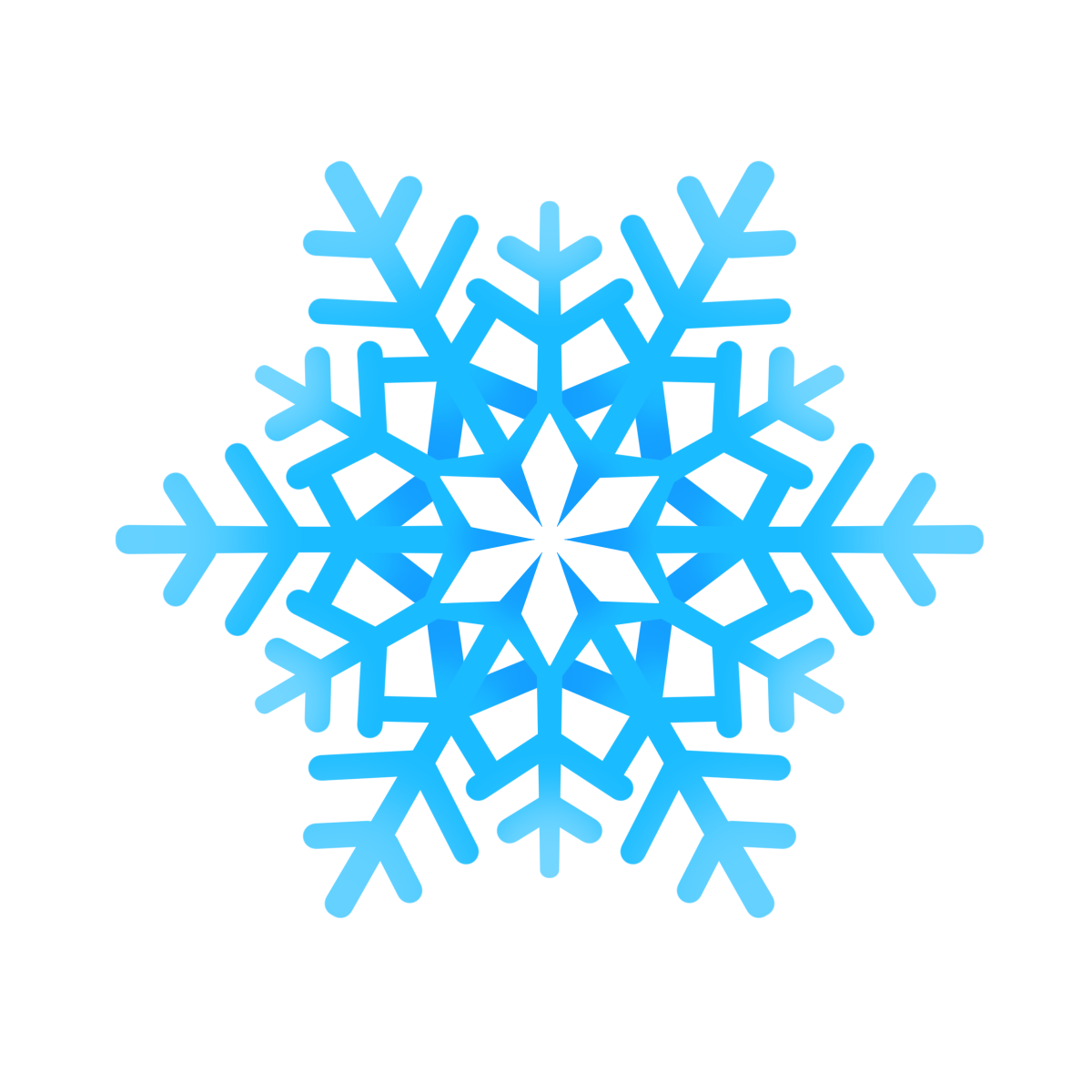}}} 
\def\mylogonarrow{\resizebox{0.5cm}{!}{\includegraphics{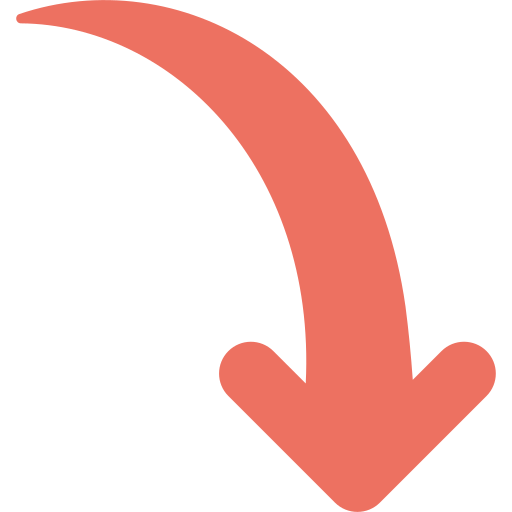}}} 
\newlength{\hseg}
\usepackage{graphicx} 
\usepackage{scalerel} 
\usepackage{hyperref}
\definecolor{reasoning-side}{RGB}{167,210,165}
\definecolor{reasoning-inside}{RGB}{206,222,186}
\definecolor{lang-side}{RGB}{159,164,248}
\definecolor{lang-inside}{RGB}{217,217,252}
\usepackage{tikz}
\newcommand{\inlinecolorboxone}{
    \tikz[baseline=-0.1cm]{
        \node[draw=lang-side, fill=lang-inside, minimum width=0.4cm, minimum height=0.1cm, line width=0.5pt, rounded corners=1pt] (char1) {};
    }
}
\newcommand{\inlinecolorboxtwo}{
    \tikz[baseline=-0.1cm]{
        \node[draw=reasoning-side, fill=reasoning-inside, minimum width=0.4cm, minimum height=0.1cm, line width=0.5pt, rounded corners=1pt] (char2) {};
    }
}
\usepackage{xcolor}
\usepackage{pgfplots}
\definecolor{tiffanyblue}{RGB}{129,216,208}
\definecolor{bangdiblue}{RGB}{0,149,182}
\definecolor{kleinblue}{RGB}{0,47,167}
\definecolor{purple}{RGB}{138,43,226}
\usepackage{amssymb}
\usetikzlibrary{shapes}
\usepackage{circledtext}
\usepackage{graphicx} 
\usetikzlibrary{arrows,decorations.pathmorphing,backgrounds,positioning,fit,petri}
\usetikzlibrary{arrows.meta,fit,shapes.arrows}
\usepackage{caption} 
\usepackage{xcolor}
\usepackage{pgfplots}
\usepackage{tikz}
\usepackage{subfig}
\usetikzlibrary{positioning,shapes,arrows,shadows,patterns}
\usetikzlibrary{calc}
\usepackage{tkz-kiviat,pgfplots}
\pgfplotsset{compat=newest}
\usepackage{xspace}
\newcommand{\ours}{\textsc{SLam}\xspace}
\newcommand{\splayers}{\textit{multilingualism-handling layers}\xspace}
\newcommand{\eg}{\emph{e.g.}}

\title{\ours: Towards Efficient Multilingual Reasoning via \\ Selective Language Alignment
}

\author{Yuchun Fan\textsuperscript{1}, Yongyu Mu\textsuperscript{1}, Yilin Wang\textsuperscript{1}, Lei Huang\textsuperscript{3}, Junhao Ruan\textsuperscript{1}, Bei Li\textsuperscript{4}, \\
{\bf Tong Xiao\textsuperscript{1,2}\thanks{\xspace\xspace Corresponding author.}, Shujian Huang\textsuperscript{5}, Xiaocheng Feng\textsuperscript{3}, \and Jingbo Zhu\textsuperscript{1,2}} \\
	\textsuperscript{1}NLP Lab, School of Computer Science and Engineering, Northeastern University, Shenyang, China\\
	\textsuperscript{2}NiuTrans Research, Shenyang, China\\
        \textsuperscript{3}Harbin Institute of Technology, Harbin, China
        \textsuperscript{4}Meituan Inc.\\
        \textsuperscript{5}National Key Laboratory for Novel Software Technology, Nanjing University\\
	\ttfamily{yuchunfan\_neu@outlook.com} \ttfamily{\{xiaotong,zhujingbo\}@mail.neu.edu.cn}
}

\begin{document}
\maketitle
\begin{abstract}
Despite the significant improvements achieved by large language models (LLMs) in English reasoning tasks, these models continue to struggle with multilingual reasoning. Recent studies leverage a full-parameter and two-stage training paradigm to teach models to first understand non-English questions and then reason. However, this method suffers from both substantial computational resource computing and catastrophic forgetting. The fundamental cause is that, with the primary goal of enhancing multilingual comprehension, an excessive number of irrelevant layers and parameters are tuned during the first stage. Given our findings that the representation learning of languages is merely conducted in lower-level layers, we propose an efficient multilingual reasoning alignment approach that precisely identifies and fine-tunes the layers responsible for handling multilingualism. Experimental results show that our method, \ours, only tunes $6$ layers' feed-forward sub-layers including $6.5-8\%$ of all parameters within 7B and 13B LLMs, achieving superior average performance than all strong baselines across $10$ languages. Meanwhile, \ours only involves one training stage, reducing training time by $4.1-11.9\times$ compared to the two-stage method\footnote{The project will be available at: \url{https://github.com/fmm170/SLAM}}.
\end{abstract}

\section{Introduction}
Large language models (LLMs) \citep{touvron2023llama2, openai2023gpt4} have demonstrated significant advancements in reasoning abilities \citep{huang2023towards}. However, these improvements are primarily focused on English, leading to inferior performance in non-English scenarios, especially in low-resource languages \citep{MathOctopus3}. 
As the demand for deploying LLMs in multilingual environments increases \citep{multilingual-survey1}, recent years have witnessed a growing interest in multilingual reasoning alignment, which aims to bridge the gap between the non-English reasoning abilities of LLMs and those in English \citep{MathOctopus3,QAlign12,MAPO13}.

\begin{figure}[t!]
    \centering
    \input{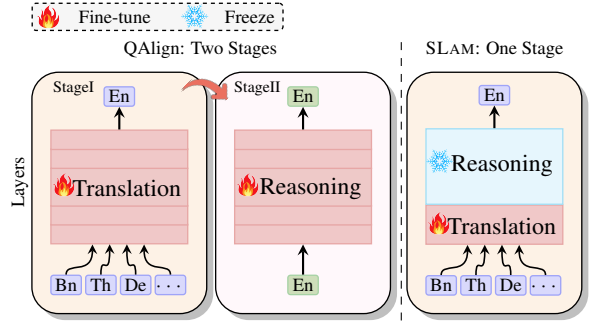}
    \caption{The comparison of QAlign with \ours (Ours). \protect\inlinecolorboxone\ and \protect\inlinecolorboxtwo\ denote the translation data and reasoning data, respectively. Unlike the traditional two-stage training approach that fully trains all parameters, \ours selectively trains only the lower-level layers responsible for multilingual comprehension in one stage.}
    \label{fig:Framework}
\end{figure}

Early work \citep{MathOctopus3} along this line of research directly fine-tunes models on multilingual mathematical question-answer pairs synthesized via machine translation.
However, the reasoning answers are too complex for accurate translation, potentially compromising the performance of fine-tuned models.
Building upon this, \citet{QAlign12} introduce a two-stage learning strategy. 
It first teaches models to comprehend multilingual inputs by translating non-English questions into English counterparts and then trains them with English-only reasoning datasets to awaken their multilingual reasoning abilities.

Despite its effectiveness, notable limitations persist with the two-stage framework. On the one hand, during its training process, all parameters of LLMs are tuned to facilitate multilingual reasoning alignment, which consumes substantial computational resources and hinders applying this method in resource-constrained scenarios. Moreover, conducting full-parameter training in the first training stage can lead to catastrophic forgetting, destroying the inherent reasoning abilities within LLMs. Thus this method depends on the second training stage to awaken the model's reasoning abilities, seeming necessary but inefficient.

When considering multilingual models, one might think of capturing language-specific knowledge in certain parts of these models. \citet{Language-Specific-Neurons} has verified that only a small number of parameters in LLMs are language-dependent. In this work, we further confirm this viewpoint by examining neuron activations across different layers of LLMs. We find that lower-level layers are more involved in learning language-specific representations, which are then transformed into universal representations in higher-level layers. This suggests that it might be worth separating the learning of language-specific and universal representations in developing multilingual reasoning LLMs.

This work is motivated by a perspective of parameter-efficient fine-tuning. In response to our findings, we precisely identify and fine-tune the \splayers with translation-only data to achieve multilingual reasoning alignment in one stage. Our method, \ours, first calculates the mean squared deviation (MSD) of the numbers of neurons activated by different languages and selects the layers with higher MSD scores. Next, recognizing that feed-forward networks (FFN) store most of the multilingual knowledge \citep{MLP-knowledge1}, \ours achieves further efficiency by only training FFN sub-layers within the selected layers. Compared to the two-stage method, \ours significantly improves the efficiency in terms of training data and computational resources. Moreover, \ours merely trains models one time since freezing irrelevant parameters effectively prevents the inherent reasoning abilities from being destroyed, as illustrated by Figure \ref{fig:Framework}.

Experimental results on two multilingual mathematical reasoning benchmarks, \textsc{mGSM} \citep{mCot4} and \textsc{mSVAMP} \citep{MathOctopus3}, show that \ours outperforms strong baselines in both in-domain and out-of-domain settings, with only 8\% and 6.5\% of the parameters tuned in 7B and 13B models, respectively. Furthermore, \ours reduces the training time by $4.1\times$ and $11.9\times$ compared to the two-stage method.
Moreover, \ours can also be generalized to multilingual common sense reasoning \citep{xCSQA} and can leverage models with advanced reasoning abilities to consistently enhance multilingual reasoning performance.

\section{Background}
\paragraph{Point-wise feed-forward network.}
Multilingual reasoning requires LLMs to integrate both multilingual comprehension and reasoning abilities. Our approach builds on the findings that factual knowledge is stored in the model’s FFN sub-layers \citep{MLP-knowledge2,MLP-knowledge3}. By directly injecting language-specific knowledge into these FFN sub-layers, we can enhance the multilingual comprehension abilities of LLMs.
Typically, each layer of Transformer-based LLMs is predominantly composed of a multi-head self-attention (MHA) sub-layer followed by an FFN sub-layer. 
Concretely, let $l$ denote the length of the input sentence. Given the output $\bm{x}^{i} \in \mathbb{R}^{l \times d_\text{model}}$ from the MHA sub-layer at the $i$-th layer, the computation within the FFN sub-layer can be formulated as follows:
\begin{equation}
  \label{eq:ffn}
  \text{FFN} (\bm{x}^{i}) = f(\bm{x}^{i} \cdot \bm{W}_\text{up}^{i}) \cdot \bm{W}_\text{down}^{i},
\end{equation}
where $\bm{W}_\text{up}^{i} \in \mathbb{R}^{d_\text{model} \times d_\text{inter}}$, and $\bm{W}_\text{down}^{i} \in \mathbb{R}^{d_\text{inter} \times d_\text{model}}$ are the mapping matrices. 
The $d_\text{model}$ denotes the input dimension, $d_\text{inter}$ refers to the intermediate hidden dimension, and $f(\cdot)$ represents a non-linear activation function.

\paragraph{Neuron activation.}
Recent work \citep{MLP-knowledge2,Multilingual-myy} reveals that language-specific neurons in the FFN sub-layers significantly influence how LLMs process multilingual languages. As a new variant of the activation function, SwiGLU \citep{SwiGLU} is widely used in current LLMs \citep{touvron2023llama2}. Thus, Equation (\ref{eq:ffn}) can further be decomposed as follows:
\begin{equation}
  \label{eq:neuron}
  \text{FFN} (\bm{x}^{i}) =  \left[f(\bm{W}_\text{gate}^{i}(\bm{x}^{i})) \otimes \bm{W}_\text{up}^{i}(\bm{x}^{i}) \right] \cdot \bm{W}_\text{down}^{i},
\end{equation}
where $\bm{W}_\text{gate}^{i} \in \mathbb{R}^{d_\text{model} \times d_\text{inter}}$ is the mapping metric introduced by SwiGLU. A neuron is defined as a single column in $\bm{W}_\text{up}^{i}$, thereby an FFN sub-layer within one layer consists of $d_\text{inter}$ neurons. 
In our work, a neuron in the $i$-th FFN sub-layer is considered activated if the value of the element in $f(\bm{W}_\text{gate}^{i}(\bm{x}^{i}))$ exceeds zero \citep{Language-Specific-Neurons}.

\section{Methodology}
 In this section, we start with a preliminary study of examining neuron activations across different layers during the multilingual reasoning process, which motivates our approach to achieving efficient multilingual reasoning alignment.
\subsection{Preliminary Study}
We examine neuron activations across different layers from two perspectives: (1) the number of neurons activated by different languages, and (2) the overlap ratio of activated neurons between non-English languages and English. We first sample $n$ questions, each expressed in 10 different languages. Let $lang$ denote the language of the input sentence.
\begin{figure}[t!]
    \centering
    \definecolor{upurple}{RGB}{155,89,182}
\definecolor{ured}{RGB}{231,76,60}
\definecolor{udark}{RGB}{77,153,77}
\definecolor{udpdark}{HTML}{85827a}
\definecolor{usemidark}{HTML}{8c564b}
\definecolor{ublue}{RGB}{52,152,219}
\definecolor{udpblue}{HTML}{0419fb}
\definecolor{usemiblue}{HTML}{17becf}
\definecolor{uorange}{HTML}{ffcc00}
\definecolor{udporange}{HTML}{bcbd22}
\tikzset{global scale/.style={
    scale=#1,
    every node/.append style={scale=#1}
  }
}

\pgfplotsset{
    width=0.685\textwidth,
   height=0.25\textheight,
   grid=major,
   major grid style={dotted},
   enlarge y limits={upper,value=0.05},
   legend style={
      fill,
      at={(0.50,16.5em)},
      legend columns=5,
      legend cell align=left,
      anchor=south
      },
   }
\begin{tikzpicture}
\useasboundingbox (13em,-1em) rectangle (3.5em,10em);
\begin{axis}[
global scale = 0.63,
   at={(1em,1em)},
    legend entries={Bn,Th,Sw,De,Fr,Zh,Ja,Ru,Es,En},
    ymajorgrids=true,
    xmajorgrids=true,
    legend style={draw=none,
    line width=1pt,
    at={(10.5em,12.5em)},
    legend cell align=left,
    column sep=3pt,
    yshift=-5pt,
    anchor=south
    },
    enlarge x limits={abs=1},
   xtick={1,2,3,4,5,6,7,8,9,10,11},
   xticklabels={1,2,3,4,5,6,7,8,9,10,11},
   ymin=0.25, ymax=0.55,
   ytick={0.25,0.3,0.35,0.4,0.45,0.5,0.55},
   yticklabels={0.25,0.30,0.35,0.40,0.45,0.50,0.55},
   yticklabel style={/pgf/number format/fixed,/pgf/number format/fixed zerofill,/pgf/number format/precision=2,rotate=0}
   ]

   \addplot[sharp plot,udpblue!40,smooth,thick,line width=0.5pt,mark=pentagon*,mark size=2pt,thick,mark options={fill=udpblue!40,draw=udpblue!55,line width=0.5pt}] plot coordinates{
      (1,0.499)(2,0.437) (3,0.410)
      (4,0.406) (5,0.394) (6,0.362)
      (7,0.318) (8,0.300) (9,0.293)
      (10,0.304) (11,0.310)
      };
   \addplot[sharp plot,usemiblue!65,smooth,thick,line width=0.5pt,mark=triangle*,mark size=2pt,thick,mark options={fill=usemiblue!65,draw=usemiblue!75,line width=0.5pt}] plot coordinates{
      (1,0.518) (2,0.468) (3,0.431)
      (4,0.422) (5,0.409) (6,0.372)
      (7,0.324) (8,0.302) (9,0.293)
      (10,0.308) (11,0.315)
      };
   \addplot[sharp plot,ublue!65,smooth,thick,line width=0.5pt,mark=square*,mark size=1.6pt,thick,mark options={fill=ublue!65,draw=ublue!75,line width=0.5pt}] plot coordinates{
      (1,0.377)(2,0.367) (3,0.372)
      (4,0.356) (5,0.348) (6,0.329)
      (7,0.292) (8,0.277) (9,0.285)
      (10,0.291) (11,0.298)
      };
   \addplot[sharp plot,usemidark!65,smooth,thick,line width=0.5pt,mark=pentagon*,,mark size=2pt,thick,mark options={fill=usemidark!65,draw=usemidark!80,line width=0.5pt}] plot coordinates {
      (1,0.401)(2,0.369) (3,0.370)
      (4,0.356) (5,0.321) (6,0.296)
      (7,0.275) (8,0.277) (9,0.291)
      (10,0.299) (11,0.308)
      };
   \addplot[sharp plot,udpdark!65,smooth,thick,line width=0.5pt,mark=triangle*,,mark size=2pt,thick,mark options={fill=udpdark!65,draw=udpdark!75,line width=0.5pt}] plot coordinates {
      (1,0.386)(2,0.354) (3,0.369)
      (4,0.348) (5,0.318) (6,0.295)
      (7,0.274) (8,0.276) (9,0.290)
      (10,0.299) (11,0.308)
      };
   \addplot[sharp plot,uorange!65,smooth,thick,line width=0.5pt,mark=square*,mark size=1.6pt,thick,mark options={fill=uorange!65,draw=uorange!75,line width=0.5pt}] plot coordinates {
      (1,0.428)(2,0.402) (3,0.404)
      (4,0.386) (5,0.355) (6,0.325)
      (7,0.289) (8,0.287) (9,0.296)
      (10,0.305) (11,0.312)
      };
   \addplot[sharp plot,orange!65,smooth,thick,line width=0.5pt,mark=pentagon*,mark size=2pt,thick,mark options={fill=orange!65,draw=orange!75,line width=0.5pt}] plot coordinates {
      (1,0.424)(2,0.397) (3,0.404)
      (4,0.383) (5,0.359) (6,0.328)
      (7,0.296) (8,0.290) (9,0.296)
      (10,0.305) (11,0.312)
      };
   \addplot[sharp plot,udporange!65,smooth,thick,line width=0.5pt,mark=triangle*,mark size=2pt,thick,mark options={fill=udporange!65,draw=udporange!75,line width=0.5pt}] plot coordinates {
      (1,0.436)(2,0.398) (3,0.396)
      (4,0.379) (5,0.340) (6,0.305)
      (7,0.282) (8,0.281) (9,0.296)
      (10,0.303) (11,0.312)
      };
   \addplot[sharp plot,udark!65,smooth,thick,line width=0.5pt,mark=square*,mark size=1.6pt,thick,mark options={fill=udark!65,draw=udark!75,line width=0.5pt}] plot coordinates {
      (1,0.381)(2,0.354) (3,0.365)
      (4,0.348) (5,0.318) (6,0.295)
      (7,0.274) (8,0.276) (9,0.290)
      (10,0.299) (11,0.308)
      };
   \addplot[sharp plot,ured!65,smooth,thick,line width=0.5pt,mark=*,mark size=1.6pt,thick,mark options={fill=ured!65,draw=ured!75,line width=0.5pt}] plot coordinates {
      (1,0.391)(2,0.347) (3,0.358)
      (4,0.339) (5,0.313) (6,0.296)
      (7,0.277) (8,0.281) (9,0.296)
      (10,0.303) (11,0.313)
      };

   \end{axis}
\node at (8.5em,-0.4em){\footnotesize Layer};

   \node  [rotate=90, align=center] at(-1.3em,4.7em){\scriptsize Neuron activation numbers};
\end{tikzpicture}
  \caption{The number of activated neurons across all languages. The count of activated neurons is normalized by the total neurons within the FFN sub-layer.}
  \label{fig:neuron-activation-numbers-MetaMath-7B}
\end{figure}
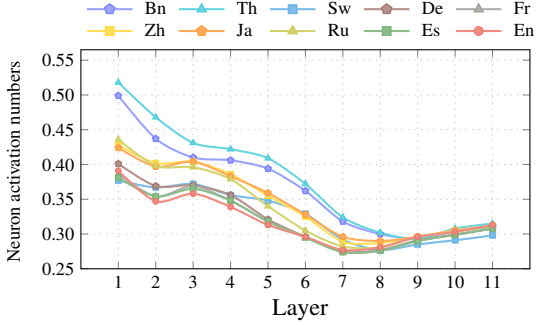
Subsequently, we calculate the count of activated neurons for all samples of a language $lang$ in the $i$-th layer, as specified by the following equation:
\begin{equation}
  \label{eq:activation}  
  A_{lang}^{i} = \mathbb{I}[f(\bm{W}_\text{gate}^{i}(x^{i}))>0 ],
\end{equation} 
where $\mathbb{I}$ is the indicator function. 
To provide a more intuitive understanding of neuronal activation, we normalize the count of activated neurons, as defined by the following equation:
\begin{equation}
  \label{eq:activated-normalization}  
  R_{lang}^{i} = \frac{A_{lang}^{i}}{d_\text{inter}}.
\end{equation} 
Then we compute the overlap ratio of activated neurons between non-English languages and English.

\begin{figure}[t!]
    \centering
    \definecolor{upurple}{RGB}{155,89,182}
\definecolor{ured}{RGB}{231,76,60}
\definecolor{udark}{RGB}{77,153,77}
\definecolor{udpdark}{HTML}{85827a}
\definecolor{usemidark}{HTML}{8c564b}
\definecolor{ublue}{RGB}{52,152,219}
\definecolor{udpblue}{HTML}{0419fb}
\definecolor{usemiblue}{HTML}{17becf}
\definecolor{uorange}{HTML}{ffcc00}
\definecolor{udporange}{HTML}{bcbd22}
\tikzset{global scale/.style={
    scale=#1,
    every node/.append style={scale=#1}
  }
}

\pgfplotsset{
    width=0.685\textwidth,
   height=0.25\textheight,
   grid=major,
   major grid style={dotted},
   symbolic x coords={1,2,3,4,5,6,7,8,9,10,11},
   enlarge y limits={upper,value=0.05},
   legend style={
      fill,
      at={(0.50,16.5em)},
      legend columns=5,
      legend cell align=left,
      anchor=south
      },
   }
\begin{tikzpicture}
\useasboundingbox (13em,-1em) rectangle (3.5em,10em);
\begin{axis}[
global scale = 0.63,
   at={(1em,1em)},
    legend entries={Bn,Th,Sw,De,Fr,Zh,Ja,Ru,Es,AVG},
    ymajorgrids=true,
    xmajorgrids=true,
    legend style={draw=none,
    line width=1pt,
    at={(10.5em,12.5em)},
    legend cell align=left,
    column sep=3pt,
    yshift=-5pt,
    anchor=south
    },
   xtick={1,2,3,4,5,6,7,8,9,10,11},
   xticklabels={1,2,3,4,5,6,7,8,9,10,11},
   ymin=0.5, ymax=0.95,
   ytick={0.5,0.55,0.6,0.65,0.7,0.75,0.8,0.85,0.9,0.95},
   yticklabels={,0.55,0.60,0.65,0.70,0.75,0.80,0.85,0.90,0.95},
   yticklabel style={/pgf/number format/fixed,/pgf/number format/fixed zerofill,/pgf/number format/precision=2,rotate=0}
   ]
   \addplot[sharp plot,udpblue!40,smooth,thick,line width=0.5pt,mark=pentagon*,mark size=2pt,thick,mark options={fill=white,draw=udpblue!40,line width=0.5pt}] plot coordinates{
      (1,0.610)(2,0.571) (3,0.571)
      (4,0.582) (5,0.593) (6,0.612)
      (7,0.647) (8,0.672) (9,0.699)
      (10,0.711) (11,0.718)
      };
   \addplot[sharp plot,usemiblue!65,smooth,thick,line width=0.5pt,mark=triangle*,mark size=2pt,thick,mark options={fill=white,draw=usemiblue!65,line width=0.5pt}] plot coordinates{
      (1,0.589)(2,0.570) (3,0.572)
      (4,0.594) (5,0.600) (6,0.616)
      (7,0.655) (8,0.674) (9,0.700)
      (10,0.714) (11,0.723)
      };
   \addplot[sharp plot,ublue!65,smooth,thick,line width=0.5pt,mark=square*,mark size=1.6pt,thick,mark options={fill=white,draw=ublue!65,line width=0.5pt}] plot coordinates{
      (1,0.622)(2,0.618) (3,0.611)
      (4,0.633) (5,0.642) (6,0.637)
      (7,0.682) (8,0.719) (9,0.737)
      (10,0.746) (11,0.758)
      };
   \addplot[sharp plot,usemidark!65,smooth,thick,line width=0.5pt,mark=pentagon*,,mark size=2pt,thick,mark options={fill=white,draw=usemidark!65,line width=0.5pt}] plot coordinates {
      (1,0.643)(2,0.646) (3,0.660)
      (4,0.709) (5,0.774) (6,0.804)
      (7,0.840) (8,0.872) (9,0.884)
      (10,0.889) (11,0.892)
      };
   \addplot[sharp plot,udpdark!65,smooth,thick,line width=0.5pt,mark=triangle*,,mark size=2pt,thick,mark options={fill=white,draw=udpdark!65,line width=0.5pt}] plot coordinates {
      (1,0.669)(2,0.660) (3,0.665)
      (4,0.730) (5,0.794) (6,0.820)
      (7,0.855) (8,0.883) (9,0.897)
      (10,0.900) (11,0.902)
      };
   \addplot[sharp plot,uorange!65,smooth,thick,line width=0.5pt,mark=square*,mark size=1.6pt,thick,mark options={fill=white,draw=uorange!65,line width=0.5pt}] plot coordinates {
      (1,0.647)(2,0.625) (3,0.655)
      (4,0.686) (5,0.719) (6,0.745)
      (7,0.791) (8,0.815) (9,0.833)
      (10,0.837) (11,0.846)
      };
   \addplot[sharp plot,orange!65,smooth,thick,line width=0.5pt,mark=pentagon*,mark size=2pt,thick,mark options={fill=white,draw=orange!65,line width=0.5pt}] plot coordinates {
      (1,0.651)(2,0.622) (3,0.650)
      (4,0.676) (5,0.708) (6,0.737)
      (7,0.775) (8,0.804) (9,0.821)
      (10,0.825) (11,0.831)
      };
   \addplot[sharp plot,udporange!65,smooth,thick,line width=0.5pt,mark=triangle*,mark size=2pt,thick,mark options={fill=white,draw=udporange!65,line width=0.5pt}] plot coordinates {
      (1,0.687)(2,0.665) (3,0.659)
      (4,0.694) (5,0.738) (6,0.774)
      (7,0.810) (8,0.844) (9,0.858)
      (10,0.864) (11,0.869)
      };
   \addplot[sharp plot,udark!65,smooth,thick,line width=0.5pt,mark=square*,mark size=1.6pt,thick,mark options={fill=white,draw=udark!65,line width=0.5pt}] plot coordinates {
      (1,0.655)(2,0.669) (3,0.677)
      (4,0.738) (5,0.797) (6,0.826)
      (7,0.859) (8,0.885) (9,0.898)
      (10,0.902) (11,0.905)
      };
   \addplot[sharp plot,red!80,smooth,thick,line width=0.5pt,mark=oplus,mark size=1.6pt,thick,mark options={fill=white,draw=red!80,line width=0.5pt}] plot coordinates {
      (1,0.641)(2,0.627) (3,0.636)
      (4,0.671) (5,0.707) (6,0.730)
      (7,0.768) (8,0.796) (9,0.814)
      (10,0.821) (11,0.828)
      };

   \end{axis}
  \draw[red!80, dashed] (15.1em, 6.24em) -- (15.1em, 1em);
  \draw[red!80, dashed] (15.1em, 6.24em) -- (1em, 6.24em);
\node at (8.5em,-0.4em){\footnotesize Layer};
\draw[->, >=stealth, semithick,black!80] (5.5em,1.4em) -- (2.3em,1.4em);
\draw[->, >=stealth, semithick,black!80] (11.45em,1.4em) -- (15.1em,1.4em);
\node at (8.5em,1.4em){\tiny Language-specific Layers};
   \node  [rotate=90] at(-1.3em,4.9em){\scriptsize Overlap Ratio};
\end{tikzpicture}
  \caption{The overlap between non-English and English activated neurons. ``AVG'' denotes the average overlap ratio among all non-English languages.}
  \label{fig:overlap-MetaMath-7B}
\end{figure}
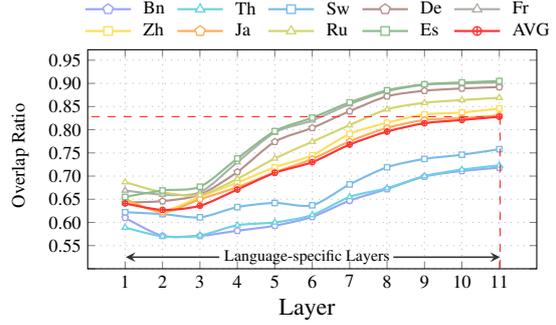

The results presented in Figures ~\ref{fig:neuron-activation-numbers-MetaMath-7B},~\ref{fig:overlap-MetaMath-7B} show that while the number of activated neurons across all languages initially decreases and then stabilizes, the neurons activated by non-English languages and English progressively overlap and finally reach a stable level.
This indicates the existence of language-specific layers, which are more involved in learning language-specific representations. The findings also align with the phenomenon found in \citet{zhao2024how}. 
Specifically, we define all layers preceding the point at which the average overlap ratio among all non-languages reaches its maximum as language-specific layers, as shown in Figure ~\ref{fig:overlap-MetaMath-7B}.
Full visualization results of all layers are presented in Figure~\ref{fig:MetaMath-7B-neurons-overlap} and Figure~\ref{fig:MetaMath-13B-neurons-overlap}, respectively.

\subsection{\ours}
In response to our findings, we formally introduce a training method to efficiently achieve multilingual reasoning alignment of LLMs in one stage. The method consists of two steps: (1) Selecting multilingualism-handling layers, and (2) Selectively supervised fine-tuning.

\paragraph{Selecting multilingualism-handling layers.}
Directly fine-tuning the language-specific layers will inevitably impair the model's inherent reasoning abilities, since reasoning abilities also persist in these layers \citep{layer-prob17}. Therefore, we design a layer selection algorithm to identify layers that are more involved in multilingual comprehension, thereby effectively balancing understanding and reasoning abilities. 
Consequently, we introduce the mean squared deviation (MSD) to precisely measure the stabilization of neuron activation across different languages. 
For each layer $i$ within the language-specific layers denoted by $\mathcal{K}$, $MSD^{i}$ is defined by the following equations: 
\begin{align}
  \label{eq:msd}  
  \mu^{i} &= \frac{1}{|L|} \sum_{lang \in L} R_{lang}^{i},  \\
  MSD^{i} &= \frac{1}{|L|} \sum_{lang \in L} (R_{lang}^{i} - \mu^{i})^2, 
\end{align}
where $L$ denotes all languages and $R_{lang}^{i}$ is calculated using  Equation (\ref{eq:activated-normalization}). 
A higher $MSD^{i}$ indicates the destabilization of neurons activated in different languages, suggesting that the layer is more actively engaged in multilingual comprehension.
The average engagement in multilingual comprehension across language-specific layers is quantified by the following equation: 
\begin{equation}
  \label{eq:theta}
\theta = \frac{1}{|\mathcal{K}|} \sum_{i \in \mathcal{K}}MSD^{i}.
\end{equation} 
We select the layers with $MSD^{i}$ exceeding threshold $\theta$ for subsequent fine-tuning, as these layers contribute most to multilingual comprehension while minimally affecting reasoning abilities.

\paragraph{Selectively supervised fine-tuning.}
Since the FFN sub-layers store the majority of the knowledge \citep{MLP-knowledge1,MLP-knowledge2}, we achieve further efficiency by only training the FFN sub-layers within the multilingualism-handling layers utilizing X-English translation data. Given non-English inputs $I_{lang}$ and their English counterparts $I_{eng}$, the optimization objective can be formulated as:
\begin{equation}
\label{eq:optimization}
\arg\min_{\varphi} \sum_{lang \in L \setminus \{\text{English}\}} -\log p_{\varphi} (I_{eng} | I_{lang}),
\end{equation}
where $\varphi$ denotes the FFN of the selected layers.

\section{Experiment Settings}
\begin{table}[t!]
    \centering

\small
\resizebox{0.49\textwidth}{!}{%
\centering
\begin{tabular}{c c c c r}

\toprule
\textbf{Task} & \textbf{Dataset} & \textbf{Usage} & \textbf{Lang} & \textbf{Number} \\
\midrule
\multirow{4}{*}{\shortstack[c]{\textbf{Multilingual}\\ \textbf{Mathematical} \\\textbf{Reasoning}}}& \textsc{MGSM-I} (Question) & Training & 10 & 57,817 \\
& \textsc{MGSM-I} (Answer) & Training & 10 & 65,968 \\
& \textsc{mGSM} & In-Domain Test& 10 & 2,500 \\
& \textsc{mSVAMP} & Out-of-Domain Test & 10 & 10,000  \\
\midrule
\addlinespace[2pt]
\multirow{3}{*}{\shortstack[c]{\textbf{Common Sense}\\\textbf{Reasoning}}}& \textsc{xCSQA-TEST} & Training & 16 & 16,110  \\
& Flores-200-DEV & Training & 16 & 14,955 \\
& \textsc{xCSQA-DEV} & Test & 16 & 16,000 \\
\bottomrule
\end{tabular}
}

  \caption{Statistics of the involved datasets. ``Lang'' denotes the number of languages covered, and ``Number'' denotes the total samples within each dataset. ``\textsc{MGSM-I}'' stands for the \textsc{MGSM8KInstruct} dataset.}
  \label{table:dataset}
\end{table}

\subsection{Baseline Models}
For a fair comparison, we compare \ours with the following strong baselines, all trained based on Llama 2  \citep{touvron2023llama2}. Detailed statistics of the training data for all baselines can be found in the Appendix~\ref{sec:appendix_training}.
\paragraph{MAmmoTH:} \citet{yue2024mammoth} collected diverse instruction datasets on MATH and directly fine-tuned the model on these datasets.
\paragraph{WizardMath:} \citet{luo2023wizardmath} leveraged reinforcement learning to train the model on two mathematical reasoning datasets GSM8K and MATH.
\paragraph{MetaMath:} \citet{Metamath10} first employed question bootstrapping to create high-quality English reasoning dataset \textsc{MetaMathQA} and then fine-tuned the model on the dataset.
\paragraph{MathOctopus:} \citet{MathOctopus3} employed supervised fine-tuning using \textsc{MSGM8KInstruct}, a multilingual reasoning dataset constructed by directly translating the data from GSM8K.
\paragraph{LangBridge:} \citet{LangBridge6} introduced an extra multilingual encoder and trained a linear layer connecting this encoder to the LLM using English data to enhance multilingual comprehension.
\paragraph{QAlign:} \citet{QAlign12} employed a two-stage training strategy, where the model first learns to translate non-English questions into English and then unlocks multilingual reasoning abilities using the English reasoning data \textsc{MetaMathQA}.

\subsection{Experimental Details}
We constructed X-English translation data from \textsc{MGSM8KInstruct} \citep{MathOctopus3} as training data. 
For \textsc{mGSM} \citep{mCot4} and \textsc{mSVAMP} \cite{MathOctopus3}, we trained only the FFN within the first six layers of the model. Considering that QAlign and LangBridge are either trained on \textsc{MetaMathQA} or built upon MetaMath, we implemented \ours on the MetaMath model to ensure a fair comparison. 
During inference, we adopted the settings from \citet{Metamath10}. 
For more details, please refer to Appendix~\ref{sec:appendix_training}.

\subsection{Evaluation Datasets}
We utilized \textsc{mGSM} and \textsc{mSVAMP} as in-domain and out-of-domain test sets to evaluate the multilingual mathematical reasoning abilities of LLMs.
Data statistics are reported in Table~\ref{table:dataset}.

\subsection{Evaluation Metrics}
Following \citet{QAlign12}, our evaluation primarily focuses on two key dimensions: Accuracy and Prediction Consistency Ratio.

\begin{table*}[t!]
    
\centering
\resizebox{1
\textwidth}{!}{%
  \begin{tabular}{l|c|c|cccccccccc|c}
    \toprule
    \multirow{2}{*}{\textbf{Model}} & 
    \multirow{2}{*}{\shortstack{\\ \textbf{Training} \\ \textbf{Cost}}} &
     \multirow{2}{*}{\shortstack{\\ \textbf{Trained} \\ \textbf{Param.}}} & 
    \multirow{2}{*}{\textbf{Bn}} & \multirow{2}{*}{\textbf{Th}} & \multirow{2}{*}{\textbf{Sw}} & \multirow{2}{*}{\textbf{Ja}} & 
    \multirow{2}{*}{\textbf{Zh}} & \multirow{2}{*}{\textbf{De}} & \multirow{2}{*}{\textbf{Fr}} & \multirow{2}{*}{\textbf{Ru}} & 
    \multirow{2}{*}{\textbf{Es}} & \multirow{2}{*}{\textbf{En}} & \multirow{2}{*}{\textbf{Avg.}} \\
    & & & & & & & & & & & & & \\

    \midrule

     \multicolumn{14}{c}{\multirow{1}{*}{\textit{7B Models}}} \\

    \midrule
 MAmmoTH$^\dagger$ & 2.7$\times$ &100.0\% & \phantom{0}3.6 &  \phantom{0}4.8 &\phantom{0}2.4 & 10.8 & 17.2 & 33.2 & 32.8 & 26.0 & 32.4 & 49.6 &21.3 \\
  WizardMath$^\dagger$ & 0.9$\times$ &  100.0\% &\phantom{0}2.0 &  \phantom{0}4.0 &\phantom{0}3.4 & 24.0 & 22.4 & 30.4 & 30.4 & 30.8 & 34.8 & 47.6 &23.0 \\
MetaMath$^\dagger$ & 3.5$\times$ &100.0\%& \phantom{0}7.6  & \phantom{0}5.6  & \phantom{0}5.2  & 34.0 & 45.2 & 54.0 & \textbf{56.8} & 51.6 & 58.8 & 65.5 & 38.4 \\
 MathOctopus$^\dagger$ & 0.8$\times$ & 100.0\% & 28.8 & 34.4 & 39.2 & 36.0 & 38.4 & 44.8 & 43.6 & 39.6 & 42.4 & 52.4 & 40.0 \\
QAlign$^\dagger$  & 4.1$\times$ & 100.0\% & \textbf{32.4} & 39.6 & \textbf{40.4} & 44.0 & \textbf{48.4} & \textbf{54.8} & \textbf{56.8} & 52.4 & \textbf{59.6} & \textbf{68.0} & \textbf{49.6} \\
\hspace{3mm}+LoRA (r=256) & 2.6$\times$ & \phantom{0}\phantom{0}8.7\%& \phantom{0}6.0 & \phantom{0}6.8 & \phantom{0}6.0 & 22.0 & 22.0 & 28.0 & 32.8 & 30.4 & 29.6 & 38.8 & 22.2 \\
    \midrule
\textbf{\ours}&1.0$\times$& \phantom{0}\phantom{0}8.0\% & 32.0 & \textbf{44.0} & 40.0 & \textbf{46.0} & \textbf{48.4} & 54.0 & 55.2 & \textbf{54.8} & 56.8 & 64.8 & \textbf{49.6} \\
    \midrule
   \multicolumn{14}{c}{\multirow{1}{*}{\textit{13B Models}}} \\
    \midrule
 MAmmoTH$^\dagger$ &7.3$\times$ & 100.0\%&\phantom{0}3.6 &  \phantom{0}5.2 &\phantom{0}1.6 & 19.2 & 31.2 & 45.6 & 39.6 & 36.8 & 50.0 & 56.4 &28.9 \\
  WizardMath$^\dagger$ & 2.5$\times$ & 100.0\%&\phantom{0}6.4 &  \phantom{0}5.6 &\phantom{0}5.6 & 22.0 & 28.0 & 40.4 & 42.0 & 34.4 & 45.6 & 52.8 &28.3 \\
MetaMath$^\dagger$ & 10.0$\times$\phantom{0} &100.0\% & 12.4 & 11.2  & \phantom{0}6.4  & 42.0 & 46.0 & \textbf{64.0} & 62.4 & 61.6 & 64.8 & 68.4 & 43.9 \\
MathOctopus$^\dagger$ & 2.0$\times$ &100.0\% & 35.2 & 46.8 & 42.8 & 43.2 & 48.8 & 44.4 & 48.4 & 47.6 & 48.0 & 53.2 & 45.8 \\
QAlign$^\dagger$ & 11.9$\times$\phantom{0} &100.0\% & 38.4 & \textbf{49.6} & 46.0 & 52.4 & \textbf{59.2} & 62.0 & 62.4 & \textbf{64.4} & 67.2 & 69.2 & 57.1 \\
\hspace{3mm}+LoRA (r=256) & 3.1$\times$ & \phantom{0}\phantom{0}7.1\% &  12.8 &15.2 & 10.8 & 35.6 & 34.4 & 46.0 & 44.4 & 40.8 &47.2 &54.8  &34.2 \\
    \midrule
\textbf{\ours} & 1.0$\times$ & \phantom{0}\phantom{0}6.5\% & \textbf{45.6} & 47.6 & \textbf{46.4} & \textbf{54.0} & 58.8 & 62.8 & \textbf{65.2} & \textbf{64.4} & \textbf{67.6} & \textbf{71.2} & \textbf{58.3} \\
    \bottomrule
  \end{tabular}

}

    \caption{The accuracy (\%) on the in-domain \textsc{mGSM} test sets. ``Avg.'' denotes the average multilingual performance and the highest score among systems of the same size are highlighted in \textbf{bold}. ``Training Cost'' denotes the time required for training models. ``Trained Param.'' indicates the proportion of trainable parameters to the model's total parameters. Results marked with $^\dagger$ come from \citet{QAlign12}.}
     \label{tab:MGSM8Kscores}
\end{table*}

\paragraph{Accuracy.}
Following \citet{Metamath10}, accuracy is calculated by comparing the last numerical value in the response to the gold answer. Higher accuracy indicates stronger reasoning ability.
\paragraph{Prediction Consistency Ratio (PCR).}
Assuming $M$ and $N$ are sets of questions correctly answered in English and non-English languages respectively. PCR is calculated as: $ \frac{|M \cap N|}{|M|}$. Higher PCR denotes the model's non-English reasoning ability is closer to its English reasoning ability. 

\section{Experimental Results}

\subsection{Main Results}

\paragraph{\ours effectively achieves multilingual reasoning alignment.} We present the results on \textsc{mGSM} in Table~\ref{tab:MGSM8Kscores}, which demonstrates that \ours outperforms all strong baselines in in-domain settings. Specifically, \ours achieves comparable performance with QAlign in 7B models and surpasses all baselines in 13B models, with an average accuracy improvement of 2.1\%. Notably, compared to MetaMath, \ours exhibits substantial improvements, achieving increases of 29.2\% and 32.8\% in the 7B and 13B models, by selectively fine-tuning multilingualism-handling layers in MetaMath using translation data. Furthermore, as shown in Figure~\ref{fig:PCR} (a), \ours demonstrates higher answer consistency. This highlights that directly enhancing multilingual comprehension at specific lower-level layers can effectively improve the multilingual reasoning abilities of LLMs.

\paragraph{\ours shows significant out-of-domain generalization.} To further validate the generalization ability of \ours, we evaluate it on the out-of-domain test sets, \textsc{mSVAMP}. As shown in Table~\ref{tab:MSVAMPscores}, \ours demonstrates robust generalization compared with all baselines, exhibiting an average accuracy improvement of 5.8\% and 0.2\% in the 7B and 13B models, respectively. Notably, \ours-7B shows substantial performance gains across all 10 languages. Moreover, as shown in Figure~\ref{fig:PCR} (b), \ours also enhances answer consistency across all non-English languages. This superior generalization is attributed to the modest number of trainable parameters in \ours. Unlike the baselines, which undergo fine-tuning all parameters on the in-domain dataset and potentially suffer from overfitting, \ours only fine-tunes partial parameters, significantly reducing the impact of domain shifts.

\begin{table}[!t]
    \centering
    \small
\resizebox{0.49\textwidth}{!}{%
\centering
\begin{tabular}{lccccc}
    \toprule
    \multirow{2}{*}{\textbf{Models}} & \multirow{2}{*}{\shortstack[c]{\textbf{Extra} \\ \textbf{Param.}}} & \multirow{2}{*}{\shortstack[c]{\textbf{Training} \\ \textbf{Cost}}} &\multirow{2}{*}{\textbf{\textsc{mGSM}}} & \multirow{2}{*}{\textbf{\textsc{mSVAMP}}}\\
    & & & &\\
    \midrule
    {MetaMath-7B} & - & 10.0$\times$\phantom{0} & 38.4 & 46.2 \\
\hspace{6mm}+\textbf{LB}$^\dagger$-9B & 2B & 0.3$\times$ & 48.8 & 52.0 \\
\hspace{6mm}+\textbf{\ours} & 0B & 1.0$\times$ & \textbf{49.6} & \textbf{60.5} \\
    \midrule
MetaMath-13B & - & 10.0$\times$\phantom{0} & 43.9 & 51.8 \\
\hspace{6mm}+\textbf{LB}$^\dagger$-20B & 7B & 0.3$\times$ & 55.8 & 57.9 \\
\hspace{6mm}+\textbf{\ours} & 0B & 1.0$\times$ & \textbf{58.3} & \textbf{62.7} \\
    \bottomrule
\end{tabular}
}
  \caption{The average accuracy of LangBridge 9B/20B models on the \textsc{mGSM} and \textsc{mSVAMP} test sets. Results marked with $^\dagger$ come from \citet{LangBridge6}.}
  \label{table:LangBridge-scores}
\end{table}

\begin{table*}[t!]
\centering
\resizebox{1\textwidth}{!}{%
  \begin{tabular}{l|c|c|cccccccccc|c}
    \toprule
    \multirow{2}{*}{{\textbf{Model}}} & \multirow{2}{*}{\shortstack{\\ \textbf{Training} \\ \textbf{Cost}}} &\multirow{2}{*}{\shortstack[c]{\textbf{Trained} \\ \textbf{Param.} }} & \multirow{2}{*}{\textbf{Bn}} & \multirow{2}{*}{\textbf{Th}} & \multirow{2}{*}{\textbf{Sw}} & \multirow{2}{*}{\textbf{Ja}} & \multirow{2}{*}{\textbf{Zh}} & \multirow{2}{*}{\textbf{De}} & \multirow{2}{*}{\textbf{Fr}} & \multirow{2}{*}{\textbf{Ru}} & \multirow{2}{*}{\textbf{Es}} & \multirow{2}{*}{\textbf{En}} & \multirow{2}{*}{\textbf{Avg.}} \\
    & & & & & & & & & & & & & \\
    \midrule
    \multicolumn{14}{c}{\multirow{1}{*}{\textit{7B Models}}} \\
    \midrule
    
 MAmmoTH$^\dagger$ & 2.7$\times$ &100.0\% & \phantom{0}4.3 &  \phantom{0}6.3 &\phantom{0}4.2 & 26.7 & 26.8 & 39.6 & 39.9 & 33.7 & 42.9 & 45.1 &26.3 \\
  WizardMath$^\dagger$ & 0.9$\times$ &  100.0\% &16.1 &  17.0 &10.3 & 37.9 & 36.3 & 39.2 & 37.7 & 37.4 & 44.8 & 48.5 &32.5 \\
MetaMath$^\dagger$  & 3.5$\times$ & 100.0\% & 15.0 & 17.1 & 15.4 & 51.9 & 54.4 & 60.9 & 62.2 & 59.3 & 63.3 & \textbf{65.5} & 46.2 \\
MathOctopus$^\dagger$ & 0.8$\times$ &100.0\% & 31.8 & 39.3 & 43.4 & 41.1 & 42.6 & 48.4 & 50.6 & 46.9 & 49.4 & 50.7 & 44.1 \\
QAlign$^\dagger$  & 4.1$\times$ & 100.0\% & 41.7 & 47.7 & 54.8 & 58.0 & 55.7 & 62.8 & 63.2 & 61.1 & 63.3 & 65.3 & 57.2 \\
\hspace{3mm}+LoRA (r=256) & 2.6$\times$& \phantom{0}\phantom{0}8.7\%& 19.3 & 20.1 & 15.1 & 33.0 & 32.7 & 46.2 & 47.3 & 41.9 & 47.2 & 51.4 & 35.4 \\
    \midrule
\textbf{\ours} & 1.0$\times$ & \phantom{0}\phantom{0}8.0\%& \textbf{49.1} & \textbf{50.6} & \textbf{55.4} & \textbf{60.6} & \textbf{63.1} & \textbf{64.6} & \textbf{65.4} & \textbf{64.1} & \textbf{66.6} & 65.3 & \textbf{60.5} \\

    \midrule
    \multicolumn{14}{c}{\multirow{1}{*}{\textit{13B Models}}} \\
    \midrule
 MAmmoTH$^\dagger$ & 7.3$\times$ &100.0\% & \phantom{0}5.0 &  13.7 &12.9 & 42.2 & 47.7 & 52.3 & 53.8 & 50.7 & 53.9 & 53.4 &38.6 \\
  WizardMath$^\dagger$ & 2.5$\times$ &  100.0\% &13.7 &  16.3 &12.5 & 29.5 & 37.0 & 48.7 & 49.4 & 43.8 & 49.4 & 56.3 &35.7 \\
MetaMath$^\dagger$  & 10.0$\times$\phantom{0} & 100.0\% & 20.6 & 20.5 & 19.1 & 57.0 & 58.8 & 68.4 & 68.1 & \textbf{67.5} & 68.9 & 68.9 & 51.8 \\
MathOctopus$^\dagger$ & 2.0$\times$ & 100.0\% & 35.2 & 41.2 & 46.8 & 39.2 & 52.0 & 47.2 & 48.0 & 45.6 & 53.2 & 56.4 & 46.5 \\
QAlign$^\dagger$  & 11.9$\times$\phantom{0} & 100.0\% & 49.2 & \textbf{55.5} & 55.2 & \textbf{64.3} & \textbf{63.8} & \textbf{69.5} & 68.1 & 66.4 & \textbf{66.4} & 67.6 & 62.6 \\
\hspace{3mm}+LoRA (r=256) & 3.1$\times$ & \phantom{0}\phantom{0}7.1\% & 30.2 & 34.8 & 24.2 & 49.3 & 54.6 & 59.7 & 60.4 & 57.0 & 59.5 & 63.0 & 49.3\\
    \midrule
\textbf{\ours} & 1.0$\times$ & \phantom{0}\phantom{0}6.5\% & \textbf{52.5} & 53.5 & \textbf{58.2} & 62.5 & 61.7 & 68.8 & \textbf{69.9} & 64.5 & 66.1 & \textbf{69.5} & \textbf{62.7} \\
    \bottomrule
  \end{tabular}

}

    \caption{The accuracy (\%) on the out-of-domain \textsc{mSVAMP} test sets.}
     \label{tab:MSVAMPscores}
\end{table*}

\paragraph{\ours demonstrates superior efficiency.} Rather than fine-tuning all layers, \ours selectively trains layers responsible for multilingual comprehension. As shown in Table~\ref{tab:MGSM8Kscores} and Table~\ref{tab:MSVAMPscores}, \ours fine-tunes only 8\% and 6.5\% of the parameters in the 7B and 13B models, respectively. Notably, compared with QAlign, \ours reduces training time by 4.1 $\times$ and 11.9 $\times$ in the 7B and 13B models. These results suggest that \ours not only achieves effective performance in multilingual reasoning but also exhibits superior efficiency. 
Additionally, we also compare \ours with LangBridge \citep{LangBridge6}, which only fine-tunes the linear layer that aligns the multilingual encoder to the LLMs. As indicated in Table~\ref{table:LangBridge-scores}, despite LangBridge having slightly less training time, it underperforms in both in-domain and out-of-domain multilingual reasoning performance compared with \ours. 
Furthermore, the additional multilingual encoder in LangBridge increases its parameters, thereby reducing its efficiency during inference and increasing deployment costs.

\subsection{Ablation Study}

\begin{figure}[t!]
    \centering
    \definecolor{upurple}{RGB}{155,89,182}
\definecolor{ublue}{RGB}{52,152,219}
\definecolor{ured}{RGB}{231,76,60}
\definecolor{udark}{RGB}{77,153,77}
\definecolor{ugreen}{RGB}{46,204,113}
\definecolor{c1purple}{HTML}{CE93D8}
\tikzset{global scale/.style={
    scale=#1,
    every node/.append style={scale=#1}
  },
      axis break gap/.initial=0mm
}

\pgfplotsset{
   width=0.4\textwidth,
   height=0.2\textheight,
   symbolic x coords={0,1,2,3,4,5,6,7,8,9,10,11},
   enlarge y limits={upper,value=0.05},
   grid=major,
   legend style={
      fill,
      at={(0.50,16.5em)},
      legend columns=3,
      legend cell align=left,
      anchor=south
      },
   major grid style={dotted},
   }
\begin{tikzpicture}
\begin{axis}[
global scale = 0.63,
   at={(1,1em)},
    name=bottom axis,   
    legend cell align=left,
    width=0.4\textwidth,
   height=0.1\textheight,
    xmin = 0, xmax = 11,     
    xtick={0,1,2,3,4,5,6,7,8,9,10,11},
    xticklabels={,1,2,3,4,5,6,7,8,R,A,},
    ymin = 14, ymax = 16.5,   
    ytick={14,15,16,17,18,19},  
     yticklabel style={/pgf/number format/fixed,rotate=0},
    grid = major,
    axis x line*=bottom,
]
\addplot [sharp plot,ublue,smooth,ultra thick,line width=1pt,mark=pentagon*,mark size=2.5pt,thick,mark options={fill=white,draw=ublue,line width=0.5pt}]
    coordinates {
    (1,43.32)(2,47.91)(3,48.16)(4,49.36)(5,49.4)(6,49.6)(7,47.11)(8,44.6)(9,21.8)(10,15.99)
    };

\end{axis}

\begin{axis}[
    global scale = 0.63,
    at=(bottom axis.north),
    legend entries={\textsc{mGSM8K}},
    anchor=south, yshift=\pgfkeysvalueof{/tikz/axis break gap},
    width=0.4\textwidth,
   height=0.25\textheight,
    xmin = 0, xmax = 11,     
    xtick={0,1,2,3,4,5,6,7,8,9,10,11},
    xticklabels={,1,2,3,4,5,6,7,8,R,A,},
    ymin = 35, ymax = 52,   
    ytick={37,39,41,43,45,47,49,51},  
    legend style={draw=none,
    line width=1pt,
    at={(6em,12.5em)},
    anchor=south},
    grid = major,
    title style={yshift=-1ex, text centered},  
    axis x line*=top,
     yticklabel style={/pgf/number format/fixed,rotate=0},
    xticklabel=\empty,
    after end axis/.code={
         \draw (rel axis cs:0,0) +(-2mm,-1mm) -- +(2mm,1mm)
              ++(0pt,-\pgfkeysvalueof{/tikz/axis break gap})
              +(-2mm,-1mm) -- +(2mm,1mm)
              (rel axis cs:0,0) +(0mm,0mm) -- +(0mm,0mm)
              ++(0pt,-\pgfkeysvalueof{/tikz/axis break gap})
              +(-2mm,-1mm) -- +(2mm,1mm);
              \draw (rel axis cs:0,0) +(-2mm,0mm) -- +(2mm,2mm)
              ++(0pt,-\pgfkeysvalueof{/tikz/axis break gap})
              +(-2mm,-1mm) -- +(2mm,1mm)
              (rel axis cs:0,0) +(0mm,0mm) -- +(0mm,0mm)
              ++(0pt,-\pgfkeysvalueof{/tikz/axis break gap})
              +(-2mm,-1mm) -- +(2mm,1mm);
    }]

\addplot [sharp plot,ublue,ultra thick,line width=0.5pt,mark=pentagon*,mark size=2.5pt,thick,mark options={fill=white,draw=ublue,line width=0.5pt}]
    coordinates {
    (1,43.32)(2,47.91)(3,48.16)(4,49.36)(5,49.4)(6,49.6)(7,47.11)(8,44.6)(9,38.07)(10,34.2)
    };

\end{axis}

\node at (4em,-0.5em){\footnotesize(a) End Training Layer};
\node at (13.6em,-0.5em){\footnotesize (b) End Training Layer};

\begin{axis}[
global scale = 0.63,
   at={(10em,1em)},
    name=bottom axis,   
    legend cell align=left,
    width=0.4\textwidth,
   height=0.1\textheight,
    xmin = 0, xmax = 11,     
    xtick={0,1,2,3,4,5,6,7,8,9,10,11},
    xticklabels={,1,2,3,4,5,6,7,8,R,A},
    ymin = 14, ymax = 16.5,   
    ytick={14,15,16,17,18,19},       
    grid = major,
    axis x line*=bottom,
]
\addplot [sharp plot,ured,ultra thick,line width=0.5pt,mark=pentagon*,mark size=2.5pt,thick,mark options={fill=white,draw=ured,line width=0.5pt}]
    coordinates {
    (1,54.893)(2,57.9)(3,59.65)(4,60.07)(5,60.4)(6,60.48)(7,59.27)(8,58.4)(9,20.5)(10,14.73)
    };

\end{axis}

\begin{axis}[
    global scale = 0.63,
    at=(bottom axis.north),
    legend entries={\textsc{mSVAMP}},
    anchor=south, yshift=\pgfkeysvalueof{/tikz/axis break gap},
    width=0.4\textwidth,
   height=0.25\textheight,
    xmin = 0, xmax = 11,     
    xtick={0,1,2,3,4,5,6,7,8,9,10,11},
    xticklabels={,1,2,3,4,5,6,7,8,R,A,},
    ymin = 45, ymax = 62,   
    ytick={47,49,51,53,55,57,59,61},          
    grid = major,
    legend style={draw=none,
    line width=1pt,
    at={(6em,13.6em)},
    anchor=south},
    title style={yshift=-1ex, text centered},  
    axis x line*=top,
    xticklabel=\empty,
    after end axis/.code={
         \draw (rel axis cs:0,0) +(-2mm,-1mm) -- +(2mm,1mm)
              ++(0pt,-\pgfkeysvalueof{/tikz/axis break gap})
              +(-2mm,-1mm) -- +(2mm,1mm)
              (rel axis cs:0,0) +(0mm,0mm) -- +(0mm,0mm)
              ++(0pt,-\pgfkeysvalueof{/tikz/axis break gap})
              +(-2mm,-1mm) -- +(2mm,1mm);
              \draw (rel axis cs:0,0) +(-2mm,0mm) -- +(2mm,2mm)
              ++(0pt,-\pgfkeysvalueof{/tikz/axis break gap})
              +(-2mm,-1mm) -- +(2mm,1mm)
              (rel axis cs:0,0) +(0mm,0mm) -- +(0mm,0mm)
              ++(0pt,-\pgfkeysvalueof{/tikz/axis break gap})
              +(-2mm,-1mm) -- +(2mm,1mm);
    }]

\addplot [sharp plot,ured,ultra thick,line width=0.5pt,mark=pentagon*,mark size=2.5pt,thick,mark options={fill=white,draw=ured,line width=0.5pt}]
    coordinates {
    (1,54.85)(2,57.9)(3,59.65)(4,60.07)(5,60.4)(6,60.48)(7,59.27)(8,58.4)(9,49.76)(10,42.55)
    };

\end{axis}
   \node  [rotate=90] at(-1.3em,5em){\scriptsize Accuracy (\%)};
   \node  [rotate=90] at(8.6em,5em){\scriptsize Accuracy (\%)};
\end{tikzpicture}
  \caption{ 
  The average accuracy of training different layers.
  The $x$-axis, ``End Training Layer'' signifies that model training encompasses all FFN sub-layers from the first through to the specified layer. ``R'' denotes randomly selected layers, and ``A'' denotes all layers.}
  \label{fig:ACCperlayer}
\end{figure}
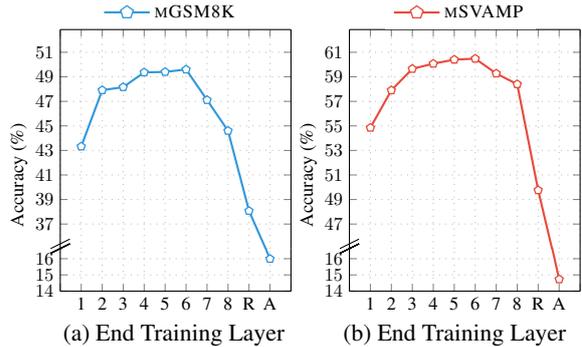

\paragraph{Training different layers.}
To evaluate the necessity of the layer selection training strategy, we conduct ablation studies by training different layers.
As shown in Figure~\ref{fig:ACCperlayer}, the layers selected by \ours achieve the best performance both in-domain and out-of-domain performance. Selecting an insufficient number of layers may lead to inadequate multilingual comprehension, while excessive layer selection will impair the model’s reasoning abilities.
This suggests that while the ability to handle multilingualism is concentrated in lower-level layers, precisely selecting layers that are more actively engaged in multilingual comprehension is crucial for effective multilingual reasoning alignment.

\paragraph{Training different sub-layers.}
To explore the role of different sub-layers within the multilingualism-handling layers, we conduct ablation studies by training only the Attention sub-layers, and both the Attention and FFN sub-layers, utilizing X-English translation data.
 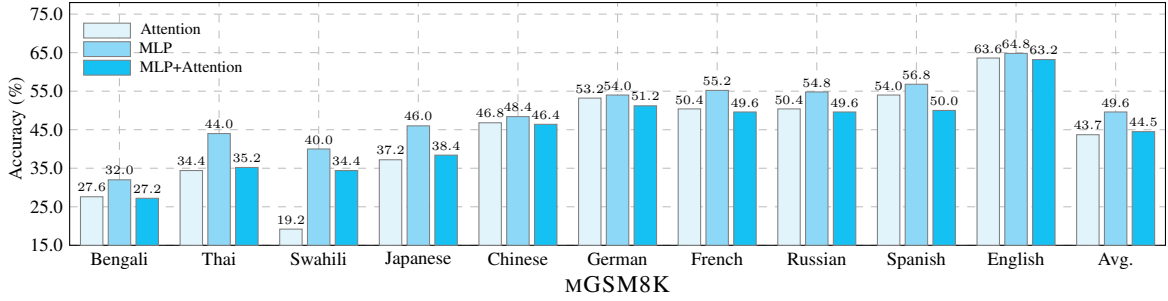
\begin{figure*}[t!]
    \centering
    \definecolor{ublue}{RGB}{52,152,219}
\definecolor{ured}{RGB}{240,100,100}
\definecolor{uorange}{RGB}{247,175,89}
\definecolor{upurple}{RGB}{148,137,250}
\definecolor{pink1}{HTML}{FFF0F6}
\definecolor{pink2}{HTML}{FCC2D7}
\definecolor{pink3}{HTML}{FAA2CA}
\begin{tikzpicture}
  \centering
    \scriptsize{
    \begin{axis}[
      at={(0,0)},
      ymajorgrids,
      xmajorgrids,
      grid style=dashed,
      legend style={at={(0.41,1)}, anchor=south west},
      legend cell align={left},
      ybar,
      enlarge x limits=0.05,
      xtick align=inside,
      height=.3\textwidth,
      width=1\textwidth,
      bar width=1.2em,
    nodes near coords, 
    nodes near coords align={vertical}, 
    nodes near coords style={font=\tiny, scale=0.8,/pgf/number format/fixed, /pgf/number format/precision=1}, 
     every node near coord/.append style={
        /pgf/number format/.cd,
            fixed,
            fixed zerofill,
            precision=1
    },
      xlabel={\footnotesize \textsc{mGSM8K}},
      xmax=10,
      xmin=0,
      symbolic x coords={0,1,2,3,4,5,6,7,8,9,10},
      legend style={cells={align=left}},
      xtick=data,
      nodes near coords align={vertical},
      ymin=15,
      ymax=78,
      ytick={15,25,35,45,55,65,75},
        yticklabels={15.0,25.0,35.0,45.0,55.0,65.0,75.0},
      xticklabels={Bengali,Thai,Swahili,Japanese,Chinese,German,French,Russian,Spanish,English,Avg.},
      xtick style={draw=none},
      ytick style={draw=none},
      ylabel style={yshift=-3em},xlabel style={yshift=0.3em,align=center},
      yticklabel style={/pgf/number format/fixed,/pgf/number format/fixed},
      legend style={draw=none,
        line width=1pt,
        at={(0.5,1.0)},
        anchor=south},
        xtick=data,
        axis on top=false,
      ]
         \addplot[fill=cyan!10,draw=gray, area legend] coordinates {(0,27.6) (1,34.4) (2,19.2) (3,37.2) (4,46.8) (5,53.2) (6,50.4) (7,50.4) (8,54) (9,63.6) (10,43.7) };
        \addplot[fill=cyan!40, draw=gray, area legend] coordinates {(0,32.0) (1,44.0) (2,40.0) (3,46.0) (4,48.4) (5,54.0) (6,55.2) (7,54.8) (8,56.8) (9,64.8) (10,49.6)  };
          \addplot[ fill=cyan!70,draw=gray, area legend] coordinates {(0,27.2) (1,35.2) (2,34.4) (3,38.4) (4,46.4) (5,51.2) (6,49.6) (7,49.6) (8,50) (9,63.2) (10,44.5) };

    \end{axis}
    } 
    \node [rectangle,draw=gray,fill=cyan!10,inner sep=2pt,minimum height=0.8em,minimum width=2.5em,font=\small,anchor=north,align=center,] (label1) at (2em,12em){};
    \node [rectangle,draw=gray,fill=cyan!40,inner sep=2pt,minimum height=0.8em,minimum width=2.5em,font=\small,anchor=north,align=center,] (label2) at (2em,11em){};
    \node [rectangle,draw=gray,,fill=cyan!70,inner sep=2pt,minimum height=0.8em,minimum width=2.5em,font=\small,anchor=north,align=center,] (label3) at (2em,10em){};
    \node [align=center] (label1_1) at ([xshift=3.4em,yshift=-0.35em]label1.north){\tiny Attention};
    \node [align=center] (label1_3) at ([xshift=4.5em,yshift=-0.35em]label3.north){\tiny MLP+Attention};
    \node [align=center] (label1_2) at ([xshift=2.6em,yshift=-0.35em]label2.north){\tiny MLP};
    \node [rotate=90]at (-2.8em,6.3em) {Accuracy (\%)};

\end{tikzpicture}
    \caption{The accuracy of training different sub-layers on \textsc{mGSM8K} test sets.}
    \label{fig:sub-layers-mgsm}
\end{figure*}
As shown in Figure~\ref{fig:sub-layers-mgsm}, training only the FFN sub-layers results in the highest average accuracy for both in-domain and out-of-domain tests (For the results of \textsc{mSVAMP}, refer to Appendix~\ref{sec:sub-layers-msvamp}). 
Specifically, training only the FFN sub-layers leads to notable improvements in low-resource languages, such as Bengali, Thai, and Swahili. 
This indicates that models can utilize the multilingual knowledge within the FFN sub-layers to enhance the comprehension of multilingual questions.

\section{Analysis}
\subsection{Scalability of \ours in Multilingual Common Sense Reasoning}
To evaluate the scalability of \ours, we extend our method to Multilingual Common Sense Reasoning (\textsc{xCSQA}) \citep{xCSQA}. More experimental details and results can be found in Appendix~\ref{sec:xcsqa_details}.
We observe that \ours improves accuracy across all languages. 
Notably, as shown in Table~\ref{table:PartXCSQA}, \ours achieves significant improvements for low-resource languages such as Swahili, Hindi, and Arabic, with increases of 22.6\%, 13.2\%, and 14.5\%, respectively.
\begin{table}[t!]
    \centering
    \centering
\resizebox{0.48\textwidth}{!}{%
  \begin{tabular}{l|cccccc|c}
    \toprule
    \textbf{Model} & \textbf{Sw} & \textbf{Hi} & \textbf{Ar} & \textbf{Zh} & \textbf{De} & \textbf{En} & \textbf{Avg.}\\
    \midrule
Llama2-mono& 22.1 & 31.9 & 31.7 & 52.3 & 58.4 & 77.6 & 45.6\\ 
\hspace{8mm}+\textbf{\ours} & 27.1 & 36.1 & 36.3 & 54.6 & 59.9 & 77.7 & 48.6\\
\midrule
Improvements& \textbf{+5.0} & \textbf{+4.2} & \textbf{+4.6} & \textbf{+2.3} & \textbf{+1.5} & \textbf{+0.1} & \textbf{+3.0}\\
    \bottomrule
\end{tabular}

}
  \caption{The accuracy (\%) on the \textsc{xCSQA} test sets.}
  \label{table:PartXCSQA}
\end{table}
\begin{table}[!t]
    \centering
    \small
\resizebox{0.49\textwidth}{!}{%
\centering
\begin{tabular}{ccccccccc}
        \toprule
        \multicolumn{2}{l}{\multirow{2}{*}[-0.1ex]{\textbf{Model}}} &\multirow{2}{*}{\shortstack[c]{\textbf{Training} \\ \textbf{Cost}}} & \multirow{2}{*}{\shortstack[c]{{\textbf{Trained}} \\ \textbf{{Param.}} }} &\multicolumn{2}{c}{\textsc{\textbf{mGSM8K}}} & \multicolumn{2}{c}{\textsc{\textbf{mSVAMP}}} \\
        && & & \multicolumn{1}{c}{\multirow{1}{*}[-0.1ex]{Non-En}} & \multicolumn{1}{c}{\multirow{1}{*}[-0.1ex]{En}} & \multicolumn{1}{c}{\multirow{1}{*}[-0.1ex]{\makecell{Non-En}}} & \multicolumn{1}{c}{\multirow{1}{*}[-0.1ex]{\makecell{En}}}\\
        \midrule
        \multicolumn{2}{l}{\multirow{1}{*}\textsc{RFT-7B}} &0.8$\times$& 100.0\% & 33.6  & \textbf{67.6} & 38.0 & \textbf{59.6} \\
        \multicolumn{2}{l}{\multirow{1}{*}{\hspace{4mm}+\textbf{\ours}}} &1.0$\times$&\phantom{0}\phantom{0}6.7\% & \textbf{44.8}  & 65.6 & \textbf{50.7} & 59.5 \\
        \midrule
        \multicolumn{2}{l}{\multirow{1}{*}{RFT-13B}}  & 2.6$\times$ &100.0\% & 38.0  & \textbf{75.2} &  45.5 & \textbf{66.3} \\
         \multicolumn{2}{l}{\multirow{1}{*}{\hspace{4mm}+\textbf{\ours}} } & 1.0$\times$ & \phantom{0}\phantom{0}5.4\%& \textbf{54.5} & 72.0 & \textbf{56.0} & 65.5\\
        \bottomrule
\end{tabular}
}
  \caption{The accuracy (\%) of RFT-MuggleMath 7B and 13B models on the \textsc{mGSM} and \textsc{mSVAMP} test sets. For accuracy of all languages, refer to Appendix~\ref{sec:RFT_full_scores}.}
  \label{table:RFT-scores}
\end{table}
The results demonstrate that our method can be effectively adapted to other multilingual reasoning tasks. 
This also suggests that the multilingual reasoning process can be decomposed into comprehend-then-reason patterns across layers. 

\subsection{Extending \ours to Other Strong Reasoning Models.}
We extend our method to the RFT-MuggleMath models \citep{RFT-Reason}, which possess stronger reasoning abilities.
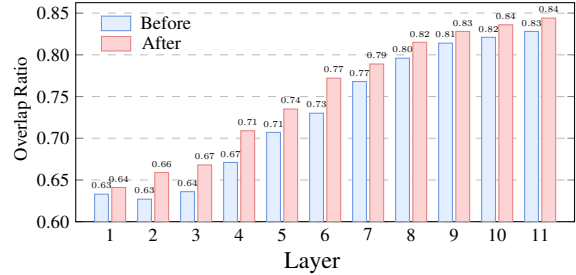
\begin{figure}[!t]
    \centering
    \definecolor{upurple}{RGB}{155,89,182}
\definecolor{ublue}{RGB}{52,152,219}
\definecolor{ured}{RGB}{231,76,60}
\definecolor{udark}{RGB}{77,153,77}
\definecolor{ugreen}{RGB}{46,204,113}
\definecolor{upink}{HTML}{fcd4d4}
\definecolor{ucyan}{HTML}{e3eeff}
\definecolor{uedgecyan}{HTML}{6d97e0}
\definecolor{uedgepink}{HTML}{cc0000}
\tikzset{global scale/.style={
    scale=#1,
    every node/.append style={scale=#1}
  }
}

\pgfplotsset{
    width=0.75\textwidth,
   height=0.25\textheight,
   symbolic x coords={1,2,3,4,5,6,7,8,9,10,11},
   enlarge y limits={upper,value=0.05},
   legend style={
      fill,
      at={(0,16.5em)},
      legend columns=2,
      legend cell align=left,
      anchor=south
      },
   }
\begin{tikzpicture}
    \begin{axis}[
    global scale = 0.63,
      at={(0.42em,1em)},
      ymajorgrids,
      grid style=dashed,
      legend style={at={(0.41,1)}, anchor=south west},
      legend cell align={left},
      ybar,
      enlarge x limits=0.08,
      xtick align=inside,
      bar width=0.75em,
    nodes near coords, 
    nodes near coords align={vertical}, 
    nodes near coords style={font=\tiny, scale=0.9,/pgf/number format/fixed, /pgf/number format/fixed zerofill, /pgf/number format/precision=2}, 
      xmax=11,
      xmin=1,
      legend style={cells={align=left}},
      xtick=data,
      nodes near coords align={vertical},
      ymin=0.60,
      ymax=0.85,
      ytick={0.6,0.65,0.7,0.75,0.8,0.85},
        yticklabels={0.60,0.65,0.70,0.75,0.80,0.85},
      xticklabels={1,2,3,4,5,6,7,8,9,10,11},
      xtick style={draw=none},
      yticklabel pos=left,
      ylabel style={yshift=-3em},xlabel style={yshift=0.3em,align=center},
      yticklabel style={/pgf/number format/fixed,/pgf/number format/fixed},
      legend style={draw=none,
        line width=1pt,
        at={(0.5,1.0)},
        anchor=south},
        xtick=data,
        axis on top=false,
      ]
         \addplot[fill=ucyan,draw=uedgecyan, area legend] coordinates { 
        (1,0.633)(2,0.627) (3,0.636)
      (4,0.671) (5,0.707) (6,0.730)
      (7,0.768) (8,0.796) (9,0.814)
      (10,0.821) (11,0.828)};
        \addplot[fill=upink, draw=uedgepink!50, area legend] coordinates { 
        (1,0.641)(2,0.659) (3,0.668)
      (4,0.709) (5,0.735) (6,0.772)
      (7,0.789) (8,0.815) (9,0.828)
      (10,0.836) (11,0.844)};

    \end{axis}
    \node [rectangle,draw=uedgecyan,fill=ucyan,inner sep=2pt,minimum height=0.5em,minimum width=1em,font=\small,anchor=north,align=center,] (label1) at (1.8em,8.1em){};
    \node [rectangle,draw=uedgepink!50,,fill=upink,inner sep=2pt,minimum height=0.5em,minimum width=1em,font=\small,anchor=north,align=center,] (label3) at (1.8em,7.4em){};
    \node [align=center] (label1_1) at ([xshift=1.55em,yshift=-0.25em]label3.north){\scriptsize After};
    \node [align=center] (label1_3) at ([xshift=1.7em,yshift=-0.25em]label1.north){\scriptsize Before};

\node at (8.5em,-0.4em){\footnotesize Layer};

   \node  [rotate=90] at(-1.4em,4.9em){\scriptsize Overlap Ratio};
   
\end{tikzpicture}
    \caption{The comparison of the overlap ratio between non-English languages and English activated neurons in the MetaMath-7B model, before and after training.}
    \label{fig:Overlap ratio}
\end{figure}
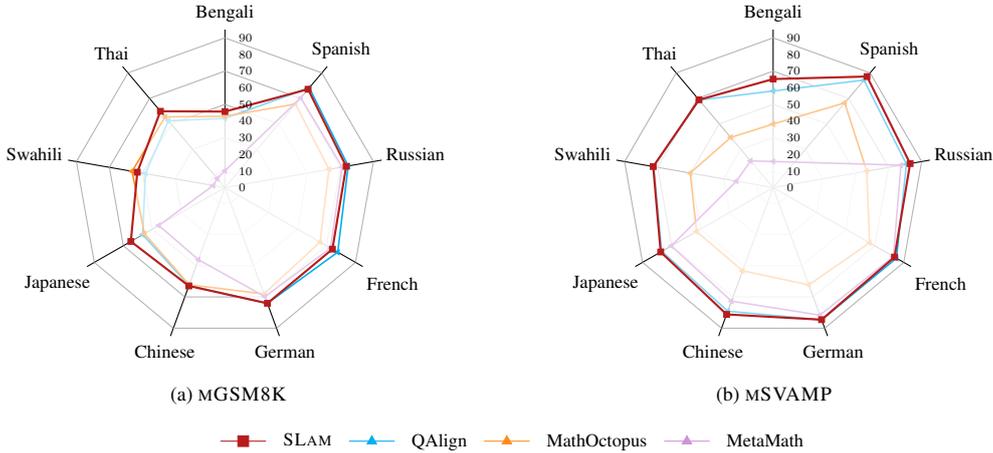
\begin{figure*}[t!]
    \centering
    \makeatletter
\def\tkz@KiviatGrad[#1](#2){%
\begingroup
\pgfkeys{/kiviatgrad/.cd,
graduation distance= 0 pt,
prefix ={},
suffix={},
unity=1,
label precision/.store in=\gradlabel@precision,
label precision=1,
zero point/.store in=\tkz@grad@zero,
zero point=0,
}
\pgfqkeys{/kiviatgrad}{#1}%
\let\tikz@label@distance@tmp\tikz@label@distance
\global\let\tikz@label@distance\tkz@kiv@grad
 \foreach \nv in {0,...,\tkz@kiv@lattice}{
 \pgfmathsetmacro{\result}{\tkz@kiv@unity*\nv+0} 
 \protected@edef\tkz@kiv@gd{%
    \tkz@kiv@prefix%
    \pgfmathprintnumber[precision=\gradlabel@precision,fixed]{\result}
    \tkz@kiv@suffix} 
    \path[/kiviatgrad/.cd,#1] (0:0)--(360/\tkz@kiv@radial*#2:\nv*\tkz@kiv@gap)
       node[label={[label distance=0.01em](360/\tkz@kiv@radial*#2):\scriptsize\tkz@kiv@gd}] {};

      }
 \let\tikz@label@distance\tikz@label@distance@tmp  
\endgroup
}%


\def\tkz@KiviatLine[#1](#2,#3){%
\begingroup
\pgfkeys{/kiviatline/.cd,
fill= {},
opacity=.5,
zero point/.store in=\tkz@line@zero,
zero point=0
}
%
%
\pgfqkeys{/kiviatline}{#1}
\ifx\tkzutil@empty\tkz@kivl@fill \else 
\path[fill=\tkz@kivl@fill,opacity=\tkz@kivl@opacity] (360/\tkz@kiv@radial*0:{(#2+\tkz@line@zero)*\tkz@kiv@gap*\tkz@kiv@step})   
\foreach \v [count=\rang from 1] in {#3}{%
 -- (360/\tkz@kiv@radial*\rang:{(\v+\tkz@line@zero)*\tkz@kiv@gap*\tkz@kiv@step}) } -- (360/\tkz@kiv@radial*0:{(#2+\tkz@line@zero)*\tkz@kiv@gap*\tkz@kiv@step});
 \fi   
\draw[#1,opacity=1,overlay] (0:{(#2+\tkz@line@zero)*\tkz@kiv@gap}) plot coordinates {(360/\tkz@kiv@radial*0:{(#2+\tkz@line@zero)*\tkz@kiv@gap*\tkz@kiv@step})}  
\foreach \v [count=\rang from 1] in {#3}{%
 -- (360/\tkz@kiv@radial*\rang:{(\v+\tkz@line@zero)*\tkz@kiv@gap*\tkz@kiv@step}) plot coordinates {(360/\tkz@kiv@radial*\rang:{(\v+\tkz@line@zero)*\tkz@kiv@gap*\tkz@kiv@step})}} -- (360/\tkz@kiv@radial*0:{(#2+\tkz@line@zero)*\tkz@kiv@gap*\tkz@kiv@step});   
\endgroup
}%

\def\tkz@KiviatDiagram[#1]#2{%

\pgfkeys{/kiviat/.cd,
gap          = .5,
lattice      = 10,
space        = .5,
step         = 1, 
label space  = 1.5
}
\pgfqkeys{/kiviat}{#1}%
\begingroup
\foreach \x [count=\rang from 1] in {#2}{%
\global\let\tkz@kiv@radial\rang}%
\foreach \x [count=\rang from 0] in {#2}{%
   \draw[/kiviatfile/radial style2]
 (0,0)--(360/\tkz@kiv@radial*\rang:\tkz@kiv@lattice*\tkz@kiv@gap+\tkz@kiv@sp);
   \path
(0,0)--(360/\tkz@kiv@radial*\rang:\tkz@kiv@lattice*\tkz@kiv@gap+\tkz@kiv@space) node[/kiviat/label style] {\x}; 

\foreach \y in {1,3,...,\tkz@kiv@lattice}{
   \draw[/kiviat/lattice style]%
     (360/\tkz@kiv@radial*\rang:\y*\tkz@kiv@gap)--%
        (360/\tkz@kiv@radial*\rang+360/\tkz@kiv@radial:\y*\tkz@kiv@gap);
}

}
\endgroup
}
\newenvironment{customlegend}[2][]{%
    \begingroup
    \pgfplots@init@cleared@structures
    \pgfplotsset{#1}%
    \begin{scope}[shift={#2}] 
}{%
    \pgfplots@createlegend
    \end{scope}
    \endgroup
}%

\def\addlegendimage{\pgfplots@addlegendimage}

\makeatother
\definecolor{c0}{HTML}{ffffff}
\definecolor{c3}{HTML}{CE93D8}
\definecolor{c4}{HTML}{ff8c1a}
\definecolor{c7}{HTML}{00aeef}
\definecolor{c8}{HTML}{B71C1C}
\definecolor{c1}{HTML}{CE93D8}
\definecolor{c2}{HTML}{ff8c1a}
\definecolor{c5}{HTML}{00aeef}
\definecolor{c6}{HTML}{B71C1C}
\tikzset{global scale/.style={
    scale=#1,
    every node/.append style={scale=#1}
  }
}
\begin{tikzpicture}[
  label distance=13em,
  global scale = 0.63,
  plot1/.style={
    semithick,
    draw=c1,
    fill=c0,
    mark=triangle*,
    mark options={
     ball color=blue, 
           color=c1,
     mark size=4pt
    },
    opacity=.5
  },
  plot2/.style={
    semithick,
    draw=c2,
    fill=c0,
    mark=triangle*,
    mark options={
      mark size=4pt,
        color=c2,
      ball color=blue
    },
    opacity=.5
  },
    plot5/.style={
    semithick,
    draw=c5,
    fill=c0,
    mark=triangle*,
    mark options={
      mark size=4pt,
        color=c5,
      ball color=blue
    },
    opacity=.5
  },
    plot6/.style={
    thick,
    draw=c6,
    fill=c0,
    mark=square*,
    mark options={
      mark size=4pt,
        color=c6,
      ball color=blue
    },
    opacity=.5
  },
  plot3/.style={
    semithick,
    draw=c3,
    fill=c0,
    mark=triangle*,
    mark options={
      mark size=4pt,
      color=c3,
      ball color=blue
    },
    opacity=.5
  },
  plot4/.style={
    semithick,
    draw=c4,
    fill=c0,
    mark=triangle*,
    mark options={
      mark size=4pt,
      color=c4,
      ball color=blue
    },
    opacity=.5
  },
    plot7/.style={
    semithick,
    draw=c7,
    fill=c0,
    mark=triangle*,
    mark options={
      mark size=4pt,
      color=c7,
      ball color=blue
    },
    opacity=.5
  },
    plot8/.style={
    thick,
    draw=c8,
    fill=c0,
    mark=square*,
    mark options={
        color=c8,
      mark size=4pt,
      ball color=c8
    },
    opacity=.5
  }
]
\useasboundingbox (-25em,-10em) rectangle (20em,7em);
\begin{scope}[scale=0.35,local bounding box=a,shift={(-50em,0)}]

\newcommand\KivStep{0.1}
\pgfmathsetmacro\Unity{1/\KivStep}
\newcommand\zeroshift{0}

 \tkzKiviatDiagram[
   radial style2/.style ={-},
   rotate=90, 
   lattice style/.style ={black!30},
   step=\KivStep,
   gap=1,
   lattice=9,
]%
{Bengali,Thai,,,Chinese,German,,,}
\node [align=center,rotate=0]at (80:30em){Swahili};
\node [align=center,rotate=0]at (-80:30em){Russian};
\node [align=center,rotate=0]at (-40:28em){Spanish};
\node [align=center,rotate=0]at (-120:30em){French};
\node [align=center,rotate=0]at (120:30em){Japanese};

\tkzKiviatLine[
  plot7
](41.57,52.41,48.19,56.63,62.65,74.1,77.71,74.7,78.3)

\tkzKiviatLine[
  plot4
](42.86,55.46,56.3,55.46,62.18,68.07,65.55,63.03,65.55)

\tkzKiviatLine[
  plot3
](10.06,7.1,7.1,45.56,46.15,69.82,72.78,71.01,70.41)
\tkzKiviatLine[
  plot8
](45.68,59.88,53.09,64.81,62.96,74.07,74.07,73.46,77.16)
\tkzKiviatGrad[unity=\Unity, label precision=2, zero point=\zeroshift](0) 
\end{scope}
\begin{scope}[scale=0.35,local bounding box=b,shift={(35em,0)}]

\newcommand\KivStep{0.1}
\pgfmathsetmacro\Unity{1/\KivStep}
\newcommand\zeroshift{0}

 \tkzKiviatDiagram[
   radial  style2/.style ={-},
   rotate=90,
   lattice style/.style ={black!30},
   step=\KivStep,
   gap=1,
   lattice=9
]%
{Bengali,Thai,,,Chinese,German,,,}
\node [align=center,rotate=0]at (80:30em){Swahili};
\node [align=center,rotate=0]at (-80:30em){Russian};
\node [align=center,rotate=0]at (-40:28em){Spanish};
\node [align=center,rotate=0]at (-120:30em){French};
\node [align=center,rotate=0]at (120:30em){Japanese};

\tkzKiviatLine[
  plot5
](58.06,68.66,72.39,76.72,79.1,84.63,84.93,80.75,84.33)
\tkzKiviatLine[
plot2
](38.19,39.35,50.23,52.78,53.24,62.27,66.44,56.71,66.44)
\tkzKiviatLine[
plot1
](15.67,20.9,22.24,70.15,72.69,81.64,82.84,77.46)
\tkzKiviatLine[
  plot6
](65.24,68.85,72.67,77.49,81.16,84.54,83.61,83,87.14)
\tkzKiviatGrad[unity=\Unity, zero point=\zeroshift](0) 
\end{scope}
\begin{customlegend}[legend columns=-1,legend style={draw=none,column sep=1ex},legend entries={\normalsize \ours,\normalsize QAlign, \normalsize MathOctopus,\normalsize MetaMath}]{(12em,-15.5em)}
    \addlegendimage{c6,fill=c6,mark=square*,sharp plot}
    \addlegendimage{c5,fill=c5,mark=triangle*,sharp plot}
    \addlegendimage{c2,fill=c2,mark=triangle*,sharp plot}
    \addlegendimage{c1,fill=c1,,mark=triangle*,sharp plot}
    \end{customlegend}

\node [rotate=0]at (-17.3em,-11.4em) {\large (a) \textsc{mGSM8K}};
\node [rotate=0]at (12.3em,-11.4em) {\large (b) \textsc{mSVAMP}};
\end{tikzpicture}
    \vspace{2.5em} 
    \caption {PCR results across various systems on the \textsc{mGSM8K} and \textsc{mSVAMP} test sets.}
    \label{fig:PCR}
\end{figure*}
As shown in Table~\ref{table:RFT-scores}, \ours achieves significant in-domain and out-of-domain improvements, with increases of 33.3\% in the average accuracy for non-English languages in both the 7B and 13B models, while tuning only 6.7\% and 5.4\% of parameters, respectively. 
This suggests that when the models possess stronger English reasoning abilities, it provides an advantageous starting point for \ours, which leads to significant improvements in multilingual reasoning alignment.

\subsection{Comparison of Neuron Activation Before and After Training}
To facilitate comparisons, we compute the average overlap ratio across all non-English languages to represent the overlap at that layer.
As shown in Figure~\ref{fig:Overlap ratio}, the overlap ratio increases significantly after training (The comparison results of MetaMath-13B, refer to Appendix~\ref{sec:neuron-activation-13B}). This demonstrates that enhancing the overlap of activated neurons between non-English languages and English at lower-level layers can improve the model's comprehension of multilingual questions, thereby achieving better multilingual reasoning alignment.
We also visualize the representations of the final token of multilingual questions, as it is crucial for the model’s subsequent reasoning \citep{Last-token-important}.
As shown in Figure~\ref{fig:Scatter}, the semantic space becomes more unified after training, thereby facilitating the sharing of abilities across languages.

\section{Related Work}
\subsection{Multilingual Mathematical Reasoning}
Significant performance discrepancies in LLM reasoning between high-resource and low-resource languages have spurred research aimed at aligning their multilingual reasoning abilities.
Early efforts \citep{MathOctopus3, mCoT-data9} directly fine-tune models on multilingual mathematical reasoning data generated via machine translation.
Another line of research concentrates on leveraging additional components during training. These works either utilize a multilingual encoder \citep{LangBridge6, MindMerger7} to facilitate cross-lingual transfer or employ an additional translation model to construct preference signals for preference optimization \citep{MAPO13}.
Recent work \citep{QAlign12} proposes a two-stage approach where the model is first learned to translate non-English questions into English for multilingual comprehension and then trained on English reasoning data to enhance multilingual reasoning abilities.
In contrast, our study particularly focuses on achieving efficient multilingual reasoning alignment, a perspective that remains under-explored. We precisely fine-tune lower-level layers that are responsible for learning language-specific representations and leverage the model’s inherent reasoning ability to facilitate multilingual reasoning alignment. This ensures superior efficiency in one stage without the need for two-stage full parameters training or introducing additional components.

\subsection{Mechanism of Multilingual Language Processing in LLMs}
Recently, multilingual LLMs have garnered significant attention. Numerous studies attempt to explore the mechanism of LLMs in processing multiple languages. 
Recent research \citep{layer-prob17, zhao2024how} reveals that both lower and upper layers of multilingual LLMs are language-dependent.
The lower layers are designed to convert inputs from various languages into a high-resource language (\eg, English), while the upper layers perform the reverse function.
Additionally, further studies \cite{Language-Specific-Neurons, Multilingual-myy} highlight that the proficiency of LLMs in comprehending a particular language is significantly influenced by a small subset of language-specific neurons. Despite their limited number, these neurons play a crucial role in bolstering the multilingual understanding abilities in LLMs.
Aligning with this line of research, \ours further reveals that specific layers are dedicated to handling multilingualism, as evidenced by neuron activation patterns.
This advancement significantly deepens the understanding of the multilingual mechanisms in LLMs.

\section{Conclusion}
In this paper, we propose \ours for efficiently achieving multilingual reasoning alignment in LLMs. Inspired by neuron activations in language abilities, we develop an approach to precisely identify the layers mostly engaging in multilingual comprehension during multilingual reasoning. After that, we fine-tune the FFN sub-layers within the selected layer to enhance the multilingual understanding abilities of LLMs. This enables achieving multilingual reasoning alignment in one stage without compromising LLMs’ inherent reasoning abilities. The experimental results on multilingual mathematical reasoning demonstrate the effectiveness and superior efficiency of \ours. Further analysis reveals that \ours exhibits significant out-of-domain generalization and can be effectively adapted to other multilingual reasoning tasks.

\section*{Limitations}
\label{sec:limitations}
Our work presents several limitations worth noting.
First, to ensure a fair comparison with baseline models, our method primarily conducts experiments using the Llama2 series models. Future work will involve extending our experiments to additional series models to more comprehensively evaluate the generalizability of our method across diverse baseline models.
Second, while our method achieves substantial advantages in average accuracy on both in-domain and out-of-domain test sets, the degrees of alignment across different languages result in performance trade-offs. We hypothesize that this issue may be due to the imbalanced data among languages in the X-English translation dataset. In the future, we will conduct an in-depth analysis of this phenomenon.

\section*{Acknowledgements}
This work was supported in part by the National Science Foundation of China (Nos. 62276056 and U24A20334), the Natural Science Foundation of Liaoning Province of China (2022-KF-16-01), the Fundamental Research Funds for the Central Universities (Nos. N2216016 and N2316002), the Yunnan Fundamental Research Projects (No. 202401BC070021), and the Program of Introducing Talents of Discipline to Universities, Plan 111 (No.B16009).

\bibliography{custom}

\newpage
\appendix
\section{Experimental Details of Multilingual Mathematical Reasoning Task}
\label{sec:appendix_training}

\subsection{Training Details}
We utilize LLaMA-Factory\footnote{\href{https://github.com/hiyouga/LLaMA-Factory}{https://github.com/hiyouga/LLaMA-Factory}} as our training framework. The training is conducted on the \textsc{MGSM8KInstruct} training dataset. Due to the complexity of mathematical reasoning questions \citep{Guo2024prompt}, hallucinations \citep{huang2024hallucination} may arise during the translation process, leading to errors in the training data, such as repeated translations. To address this issue, we conduct data quality filtering. This process yields 57,817 question translations and 65,968 answer translations. All models are trained using NVIDIA A800 GPUs. All models are trained for 4 epochs with a batch size of 512, and the learning rate is set to 2e-5. We set the maximum token length to be 1024. We use Deepspeed stage 2 \citep{deepspeed2} to conduct multi-GPU distributed training, with training precision Bfloat16 enabled.

\subsection{Training Prompts}
As shown in Table~\ref{tab:Trans-template}, we adopt the training prompt from \citet{ALMA8}.
In the template, \{\textit{source\_lang}\} can be replaced with any of the following languages: Bengali, Thai, Swahili, Japanese, Chinese, German, French, Russian, and Spanish. The placeholder \{\textit{source sentence}\} is substituted with the multilingual mathematical reasoning question (or answer), and \{\textit{English sentence}\} is replaced with the corresponding English question (or answer) that conveys the same meaning.

\begin{table*}[!t]
    \centering
\resizebox{1.0\linewidth}{!}{
  \begin{tabular}{l|p{0.7\textwidth}}
    \toprule  
    \multirow{2}{*}{\textbf{Training Prompt}} &   Translate this from  [\{\textit{source\_lang}\}] to [\text{English}]:\texttt{\string\n}[\{\textit{source\_lang}\}]: \{\textit{source sentence}\}\texttt{\string\n}[\text{English}]: \{\textit{English sentence}\}  	\\
    \bottomrule
  \end{tabular}

}
    \caption{The prompt used to train the FFN sub-layers within the multilingualism-handling layers.}
     \label{tab:Trans-template}
\end{table*}

\subsection{Evaluation Details}
\label{sec:appendix_evaluation-prompt}
During inference, we use greedy decoding to ensure the determinism of the outputs and set the maximum generation length to 512. We adopt the inference prompt from \citet{Metamath10}. The evaluation prompt for both the \textsc{mGSM} and \textsc{mSVAMP} test sets is shown in Table~\ref{tab:prompt}.

\subsection{The Training Data of All Baselines}
\label{sec:baselines_training_data}
The number of training samples used for all baselines is presented in Table~\ref{tab:baselines-data-amount}.

\section{The results of training different sub-layers on \textsc{mSVAMP} test sets}
\label{sec:sub-layers-msvamp}
\begin{figure*}[!t]
    \definecolor{ublue}{RGB}{52,152,219}
\definecolor{ured}{RGB}{240,100,100}
\definecolor{uorange}{RGB}{247,175,89}
\definecolor{upurple}{RGB}{148,137,250}
\definecolor{pink1}{HTML}{FFF0F6}
\definecolor{pink2}{HTML}{FCC2D7}
\definecolor{pink3}{HTML}{FAA2CA}
\begin{tikzpicture}
  \centering
    \scriptsize{
    \begin{axis}[
      at={(0em,-17em)},
      ymajorgrids,
      xmajorgrids,
      grid style=dashed,
      legend style={at={(0.02,0.65)}, anchor=south west},
      legend cell align={left},
      ybar,
      enlarge x limits=0.05,
      xtick align=inside,
      height=.3\textwidth,
      width=1\textwidth,
      bar width=1.2em,
    nodes near coords, 
    nodes near coords align={vertical}, 
    nodes near coords style={font=\tiny, scale=0.8,/pgf/number format/fixed, /pgf/number format/precision=1}, 
      xlabel={\footnotesize \textsc{mSVAMP}},
      symbolic x coords={0,1,2,3,4,5,6,7,8,9,10},
      legend style={cells={align=left}},
      xtick=data,
      nodes near coords align={vertical},
      ymin=40.00,
      ymax=78.00,
      ytick={40,45,50,55,60,65,70,75,78},
    yticklabels={40.0,45.0,50.0,55.0,60.0,65.0,70.0,75.0,},
      xticklabels={Bengali,Thai,Swahili,Japanese,Chinese,German,French,Russian,Spanish,English,Avg.},
      xtick style={draw=none},
      ytick style={draw=none},
      ylabel style={yshift=-3em},xlabel style={yshift=0.3em,align=center},
      yticklabel style={/pgf/number format/fixed},
      legend style={draw=none,
        line width=1pt,
        at={(0.5,1.0)},
        anchor=south},
        xtick=data,
        axis on top=false,
      ]
         \addplot[fill=pink1!50,draw=gray, area legend] coordinates {(0,48.4) (1,51.1) (2,48.8) (3,61.3) (4,60.9) (5,67.3) (6,67.1) (7,63.3) (8,64.6) (9,69.1) (10,60.2) };
          \addplot[fill=pink2!70, draw=gray, area legend] coordinates {(0,49.1) (1,50.6) (2,55.4) (3,60.6) (4,63.1) (5,64.6) (6,65.4) (7,64.1) (8,66.6) (9,65.3) (10,60.5)  };
          \addplot[ fill=pink3!70,draw=gray, area legend] coordinates {(0,48.1) (1,51.6) (2,55.1) (3,58.4) (4,58.4) (5,63.5) (6,64.3) (7,62.3) (8,61.5) (9,63.7) (10,58.7) };

    \end{axis}
    }
    

    \node [rectangle,draw=gray,fill=pink1!50,inner sep=2pt,minimum height=0.8em,minimum width=2.5em,font=\small,anchor=north,align=center,] (label1) at (2em,-5em){};
    \node [rectangle,draw=gray,fill=pink2!70,inner sep=2pt,minimum height=0.8em,minimum width=2.5em,font=\small,anchor=north,align=center,] (label2) at (2em,-6em){};
    \node [rectangle,draw=gray,,fill=pink3!70,inner sep=2pt,minimum height=0.8em,minimum width=2.5em,font=\small,anchor=north,align=center,] (label3) at (2em,-7em){};
    
    \node [align=center] (label1_1) at ([xshift=3.4em,yshift=-0.35em]label1.north){\tiny Attention};
    \node [align=center] (label1_3) at ([xshift=4.5em,yshift=-0.35em]label3.north){\tiny MLP+Attention};
    \node [align=center] (label1_2) at ([xshift=2.6em,yshift=-0.35em]label2.north){\tiny MLP};
\node [rotate=90]at (-2.8em,-11.1em) {Accuracy (\%)};

\end{tikzpicture}
    \caption{The accuracy (\%) of training different sub-layers on \textsc{mSVAMP} test sets.}
     \label{fig:sub-layers-msvmap}
\end{figure*}
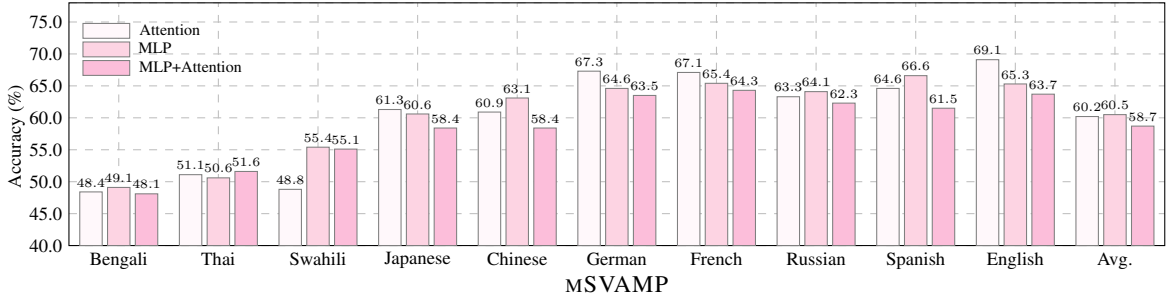

The accuracy of training only the Attention sub-layers, and both the Attention and FFN sub-layers on \textsc{mSVAMP} is shown in Figure~\ref{fig:sub-layers-msvmap}.

\section{Experimental Details of Multilingual Common Sense Reasoning Task}
\label{sec:xcsqa_details}

\subsection{Training Details}
Initially, we fine-tune the Llama2-7B base model using the English instruction dataset \textsc{xCSQA-TRAIN} to equip the model with fundamental English common sense reasoning abilities. The resulting model is named Llama2-mono. Following \citet{QAlign12}, we use the \textsc{xCSQA-TEST} datasets to construct the X-English translation data for fine-tuning. 
During the training process of \ours, we use the same layer selection approach to select the first four layers of the model as multilingualism-handling layers in \textsc{xCSQA}. Subsequently, \ours trains only the FFN sub-layers within the selected layers of the model to achieve multilingual reasoning alignment.
All models are trained for 3 epochs using NVIDIA A800 GPUs. The learning rate is set to 2e-5, with a total batch size of 512 and a maximum input length of 512.

\subsection{Training Prompts}
We use the training prompt \citep{cky-prompt} shown in Table~\ref{tab:Trans-template}. 
In the template, \{\textit{source\_lang}\} can be replaced with any of the following languages: Arabic, German, Spanish, French, Hindi, Italian, Japanese, Dutch, Polish, Portuguese, Russian, Swahili, Urdu, Vietnamese, and Chinese. The placeholder \{\textit{source sentence}\} is substituted with the multilingual common sense reasoning question, and \{\textit{English sentence}\} is replaced with the corresponding English question that conveys the same meaning.

\subsection{Evaluation Details}
Following \citet{QAlign12}, we employ \textsc{xCSQA-DEV} for evaluation. During the evaluation, we calculate accuracy by comparing the last label within brackets in the LLM-generated response with the gold answer. Specifically, we use greedy decoding to ensure the determinism of the outputs and set the maximum generation length to 512. 
The evaluation prompt is shown in Table~\ref{tab:XCSQA-prompt}.

\subsection{Languages Included in \textsc{xCSQA}}

The \textsc{xCSQA} test sets include 15 non-English languages: Arabic, German, Spanish, French, Hindi, Italian, Japanese, Dutch, Polish, Portuguese, Russian, Swahili, Urdu, Vietnamese, and Chinese. The abbreviations for these languages are as follows: ar, de, es, fr, hi, it, ja, nl, pl, pt, ru, sw, ur, vi, and zh.

\subsection{Scores for All Languages}
\label{sec:xcsqa_full_scores}
Table~\ref{tab:XCSQA} presents the scores for all 16 languages on the \textsc{xCSQA} test sets.

\begin{table*}[!h]
    \centering
\resizebox{1.0\linewidth}{!}{
  \begin{tabular}{l|p{0.7\textwidth}}
    \toprule  
\multirow{3}{*}{\textbf{Inference Prompt}} &   Below is an instruction that describes a task.\texttt{\string\n}
Write a response that appropriately completes the request.\texttt{\string\n}\texttt{\string\n}\#\#\# Instruction:\texttt{\string\n}\{\textit{query}\}\texttt{\string\n}\texttt{\string\n}\#\#\# Response: Let's think step by step.\\
    
    \bottomrule
  \end{tabular}

}

    \caption{Prompt utilized to evaluate the model on the \textsc{mGSM} and \textsc{mSVAMP} test sets.}
     \label{tab:prompt}
\end{table*}

\begin{table*}[!h]
    \centering
\resizebox{1.0\linewidth}{!}{
  \begin{tabular}{l|p{0.7\textwidth}}
    \toprule  
\multirow{3}{*}{\textbf{Inference Prompt}} &   Below is an instruction that describes a task.\texttt{\string\n}
Write a response that appropriately completes the request.\texttt{\string\n}\texttt{\string\n}\#\#\# Instruction:\texttt{\string\n}\{\textit{query}\}\texttt{\string\n}\texttt{\string\n}\#\#\# Response:\\
    
    \bottomrule
  \end{tabular}

}
    \caption{Prompt utilized to evaluate the model on the \textsc{xCSQA} test sets.}
     \label{tab:XCSQA-prompt}
\end{table*}

\begin{table*}[!h]
\centering
\resizebox{1\textwidth}{!}{%
  \begin{tabular}{l|cccccccccccccccc|cc}
    \toprule
    \textbf{Model} & \textbf{Ar} & \textbf{De} &  \textbf{Es} & \textbf{Fr} & \textbf{Hi} & \textbf{It} & \textbf{Ja} & \textbf{Nl} & \textbf{Pl} & \textbf{Pt} & \textbf{Ru} & \textbf{Sw} & \textbf{Ur} & \textbf{Vi} & \textbf{Zh} &  \textbf{En} &\textbf{Avg.}\\
    \midrule
Llama2-mono& 31.7 & 58.4 &  63.6 & 58.3 & 31.9 & 58.7 & 49.8 & 55.4 & 53.4 & 60.7 & 55.7 & 22.1 & 25.4 & 47.4 & 52.3 & 77.6 & 50.1 \\ 
\hspace{8mm}+\textbf{\ours}& 36.3 & 59.9 &  63.8 & 59.4 & 36.1 & 59.4 & 52.3 & 57.7 & 55.2 & 61.4 & 56.6 & 27.1 & 27.8 & 49.8 & 54.6 & 77.7 & 52.1 \\ 
\midrule
Improvements& \textbf{+4.6}  & \textbf{+1.5} &  \textbf{+0.2} & \textbf{+1.1} & \textbf{+4.2} & \textbf{+0.7} & \textbf{+2.5} & \textbf{+2.3} & \textbf{+1.8} & \textbf{+0.7} & \textbf{+0.9} & \textbf{+5.0} & \textbf{+2.4} & \textbf{+2.4} & \textbf{+2.3} & \textbf{+0.1} &\textbf{2.0} \\ 

    \bottomrule
  \end{tabular}

}
    \caption{The accuracy (\%) for all languages on the \textsc{xCSQA} test sets.}
     \label{tab:XCSQA}
\end{table*}

\section{Scores Across All Languages for the RFT-MuggleMath Model}
Consistent with using MetaMath as the base model, we apply the identical layer selection approach and settings for the RFT-MuggleMath model. For more details on the overlap between non-English and English activated neurons, and the number of activated neurons during the layer selection process, we present the results of all layers in Figure~\ref{fig:RFT-7B-neurons-overlap} and Figure~\ref{fig:RFT-13B-neurons-overlap}, respectively.
Specifically, we select the first five layers of the RFT-MuggleMath models as multilingualism-handling layers.
Tables~\ref{tab:RFTMGSMscores} and Tables~\ref{tab:RFTMSVAMPscores} present the accuracy for all languages on \textsc{mGSM8K} and \textsc{mSVAMP} test sets, respectively.

\label{sec:RFT_full_scores}
\begin{table*}[t!]
    
\centering
\resizebox{1
\textwidth}{!}{%
  \begin{tabular}{l|c|c|cccccccccc|c}
    \toprule
    \multirow{2}{*}{\textbf{Model}} & 
    \multirow{2}{*}{\shortstack[c]{\textbf{Training} \\ \textbf{Cost}}} &
     \multirow{2}{*}{\shortstack{\\ \textbf{Trained} \\ \textbf{Param.}}} & 
    \multirow{2}{*}{\textbf{Bn}} & \multirow{2}{*}{\textbf{Th}} & \multirow{2}{*}{\textbf{Sw}} & \multirow{2}{*}{\textbf{Ja}} & 
    \multirow{2}{*}{\textbf{Zh}} & \multirow{2}{*}{\textbf{De}} & \multirow{2}{*}{\textbf{Fr}} & \multirow{2}{*}{\textbf{Ru}} & 
    \multirow{2}{*}{\textbf{Es}} & \multirow{2}{*}{\textbf{En}} & \multirow{2}{*}{\textbf{Avg.}} \\
    & & & & & & & & & & & & & \\
    \midrule

    \multicolumn{14}{c}{\multirow{1}{*}{\textit{7B Models}}} \\

    \midrule
 RFT& 0.8$\times$ &100.0\% & \phantom{0}4.4  &\phantom{0}6.0  & \phantom{0}4.0  & 34.4 & 38.4 & 56.0 & \textbf{56.0} & 48.4 & 54.8 & \textbf{67.6} & 37.0 \\ 
    \midrule
RFT+\textbf{\ours}  & 1.0$\times$ &\phantom{0}\phantom{0}6.7\% & \textbf{27.6} & \textbf{30.0} & \textbf{33.6} & \textbf{41.6} & \textbf{42.0} & \textbf{60.4} & 55.2 & \textbf{53.2} & \textbf{60.0} & 65.6 & \textbf{46.9} \\
    \midrule
    \multicolumn{14}{c}{\multirow{1}{*}{\textit{13B Models}}} \\
    \midrule
RFT& 2.6$\times$ &100.0\% & 11.2 & \phantom{0}5.2  & \phantom{0}7.2  & 48.8 & 50.8 & \textbf{63.2} & 66.8 & \textbf{64.8} & 67.6 & \textbf{75.2} & 46.0 \\ 
    \midrule
RFT+\textbf{\ours}  & 1.0$\times$ &\phantom{0}\phantom{0}5.4\%& \textbf{42.0} & \textbf{41.6} & \textbf{42.4} & \textbf{54.0} & \textbf{51.6} &61.2 & \textbf{68.0} & 61.2 & \textbf{68.8} & 72.0 & \textbf{56.3} \\
    \bottomrule
  \end{tabular}

}

    \caption{The accuracy (\%) on the in-domain \textsc{mGSM8K} test sets of RFT-MuggleMath models.}
     \label{tab:RFTMGSMscores}
\end{table*}

\begin{table*}[t!]
\centering
\resizebox{1\textwidth}{!}{%
  \begin{tabular}{l|c|c|cccccccccc|c}
    \toprule
        \multirow{2}{*}{\textbf{Model}} & 
    \multirow{2}{*}{\shortstack[c]{\textbf{Training} \\ \textbf{Cost}}} &
     \multirow{2}{*}{\shortstack{\\ \textbf{Trained} \\ \textbf{Param.}}} & 
    \multirow{2}{*}{\textbf{Bn}} & \multirow{2}{*}{\textbf{Th}} & \multirow{2}{*}{\textbf{Sw}} & \multirow{2}{*}{\textbf{Ja}} & 
    \multirow{2}{*}{\textbf{Zh}} & \multirow{2}{*}{\textbf{De}} & \multirow{2}{*}{\textbf{Fr}} & \multirow{2}{*}{\textbf{Ru}} & 
    \multirow{2}{*}{\textbf{Es}} & \multirow{2}{*}{\textbf{En}} & \multirow{2}{*}{\textbf{Avg.}} \\
    & & & & & & & & & & & & & \\
    \midrule
     \multicolumn{14}{c}{\multirow{1}{*}{\textit{7B Models}}} \\
    \midrule
RFT& 0.8$\times$ & 100.0\% & 14.0 & 12.5 & \phantom{0}9.9  & 44.2 & 46.4 & 53.5 & 56.1 & 46.7 & 59.1 & \textbf{59.6} & 40.2 \\ 
    \midrule
RFT+\textbf{\ours} &  1.0$\times$&\phantom{0}\phantom{0}6.7\% & \textbf{40.6} & \textbf{44.2} & \textbf{44.7} & \textbf{51.4} & \textbf{50.2} & \textbf{56.8} & \textbf{58.3} & \textbf{50.3} & \textbf{59.6} &  59.5 & \textbf{51.6} \\

    \midrule
    \multicolumn{14}{c}{\multirow{1}{*}{\textit{13B Models}}} \\
    \midrule
RFT& 2.6$\times$ & 100.0\% &20.2 & 20.7 & 14.8 & 55.7 & 52.8 & \textbf{62.3} & \textbf{62.7} & 59.2 & 61.8 & \textbf{66.3} & 47.6 \\ 
    \midrule
RFT+\textbf{\ours} & 1.0$\times$& \phantom{0}\phantom{0}5.4\%& \textbf{45.3} & \textbf{46.9} & \textbf{50.3} & \textbf{58.5} & \textbf{56.1} & 61.7 & 61.8 & \textbf{60.7} & \textbf{62.5} & 65.5 & \textbf{56.9} \\
    \bottomrule
  \end{tabular}

}

    \caption{The accuracy (\%) on the out-of-domain \textsc{mSVAMP} test sets of RFT-MuggleMath models.}
     \label{tab:RFTMSVAMPscores}
\end{table*}

\section{Comparison of neuron activation before and after training of \ours-13B}
\label{sec:neuron-activation-13B}
\begin{figure}[H]
    \centering
    \definecolor{upurple}{RGB}{155,89,182}
\definecolor{ublue}{RGB}{52,152,219}
\definecolor{ured}{RGB}{231,76,60}
\definecolor{udark}{RGB}{77,153,77}
\definecolor{ugreen}{RGB}{46,204,113}
\definecolor{upink}{HTML}{fcd4d4}
\definecolor{ucyan}{HTML}{e3eeff}
\definecolor{uedgecyan}{HTML}{6d97e0}
\definecolor{uedgepink}{HTML}{cc0000}
\tikzset{global scale/.style={
    scale=#1,
    every node/.append style={scale=#1}
  }
}

\pgfplotsset{
    width=0.75\textwidth,
   height=0.25\textheight,
   symbolic x coords={1,2,3,4,5,6,7,8,9,10,11},
   enlarge y limits={upper,value=0.05},
   legend style={
      fill,
      at={(0,16.5em)},
      legend columns=2,
      legend cell align=left,
      anchor=south
      },
   }
\begin{tikzpicture}
    \begin{axis}[
    global scale = 0.63,
      at={(0.42em,1em)},
      ymajorgrids,
      grid style=dashed,
      legend style={at={(0.41,1)}, anchor=south west},
      legend cell align={left},
      ybar,
      enlarge x limits=0.08,
      xtick align=inside,
      bar width=0.75em,
    nodes near coords, 
    nodes near coords align={vertical}, 
    nodes near coords style={font=\tiny, scale=0.9,/pgf/number format/fixed, /pgf/number format/fixed zerofill, /pgf/number format/precision=2}, 
      xmax=11,
      xmin=1,
      legend style={cells={align=left}},
      xtick=data,
      nodes near coords align={vertical},
      ymin=0.60,
      ymax=0.85,
      ytick={0.6,0.65,0.7,0.75,0.8,0.85},
        yticklabels={0.60,0.65,0.70,0.75,0.80,0.85},
      xticklabels={1,2,3,4,5,6,7,8,9,10,11},
      xtick style={draw=none},
      yticklabel pos=left,
      ylabel style={yshift=-3em},xlabel style={yshift=0.3em,align=center},
      yticklabel style={/pgf/number format/fixed,/pgf/number format/fixed},
      legend style={draw=none,
        line width=1pt,
        at={(0.5,1.0)},
        anchor=south},
        xtick=data,
        axis on top=false,
      ]
         \addplot[fill=ucyan,draw=uedgecyan, area legend] coordinates { 
        (1,0.632)(2,0.625) (3,0.623)
      (4,0.668) (5,0.699) (6,0.723)
      (7,0.751) (8,0.779) (9,0.792)
      (10,0.8000) (11,0.804)};
        \addplot[fill=upink, draw=uedgepink!50, area legend] coordinates { 
        (1,0.641)(2,0.665) (3,0.659)
      (4,0.689) (5,0.725) (6,0.760)
      (7,0.773) (8,0.794) (9,0.815)
      (10,0.820) (11,0.824)};

    \end{axis}
    \node [rectangle,draw=uedgecyan,fill=ucyan,inner sep=2pt,minimum height=0.5em,minimum width=1em,font=\small,anchor=north,align=center,] (label1) at (1.8em,8.1em){};
    \node [rectangle,draw=uedgepink!50,,fill=upink,inner sep=2pt,minimum height=0.5em,minimum width=1em,font=\small,anchor=north,align=center,] (label3) at (1.8em,7.4em){};
    \node [align=center] (label1_1) at ([xshift=1.55em,yshift=-0.25em]label3.north){\scriptsize After};
    \node [align=center] (label1_3) at ([xshift=1.7em,yshift=-0.25em]label1.north){\scriptsize Before};

\node at (8.5em,-0.4em){\footnotesize Layer};

   \node  [rotate=90] at(-1.4em,4.9em){\scriptsize Overlap Ratio};
   
\end{tikzpicture}
    \caption{The comparison of the overlap ratio in the MetaMath-13B model, before and after training.}
    \label{fig:Overlap ratio-13B}
\end{figure}
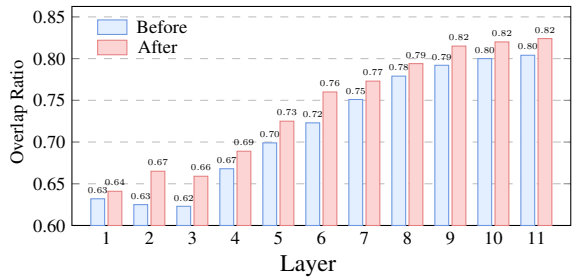

We present the comparison results in Figure~\ref{fig:Overlap ratio-13B}. 

\begin{table}[H]
    \centering
    \centering
\resizebox{0.48\textwidth}{!}{%

  \begin{tabular}{l|cc}
    \toprule
    \textbf{Model} & \textbf{Lang} & \textbf{Number} \\
    \midrule
\textbf{MAmmoTH} \citep{yue2024mammoth} & 1 &262,039 \\ 
\textbf{WizardMath} \citep{luo2023wizardmath} & 1 &96,000 \\ 
\textbf{MetaMath} \citep{Metamath10} & 1 &395,000 \\
\textbf{MathOctopus} \citep{MathOctopus3} & 10 &73,559 \\ 
\textbf{QAlign} \citep{QAlign12} & 10 &468,559 \\ 
    \bottomrule
\end{tabular}

}
    \caption{``Lang'' denotes the number of languages covered, and ``Number'' denotes the total samples used to train the models.}
    \label{tab:baselines-data-amount}
\end{table}

\begin{figure*}[t!]
    \centering
    \includegraphics[height=8cm,width=0.9\textwidth]{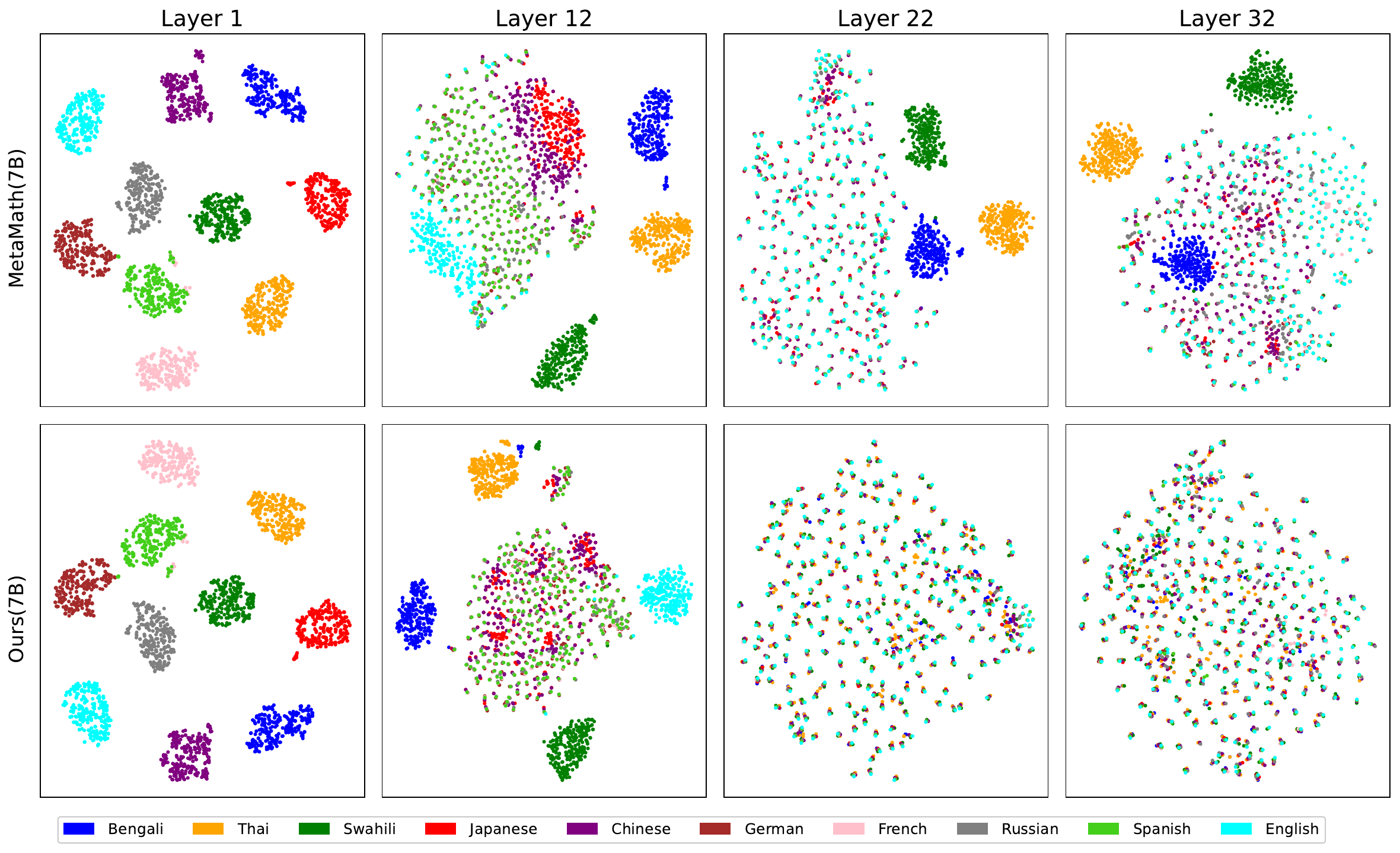}
    \caption {The visualization of the final token representations from multilingual questions is performed using T-SNE for dimension reduction. The distributions in MetaMath-7B at the 1\textsuperscript{th}, 12\textsuperscript{th}, 22\textsuperscript{th}, and 32\textsuperscript{th} layers are compared before and after training. Different colors represent different languages.}
    \label{fig:Scatter}
\end{figure*}

\begin{figure*}[t!]
    \centering
    \input{Appendix-figures/MetaMath-7B-neurons-overlap}
    \caption{The overlap between non-English and English activated neurons and the normalized number of activated neurons across all languages in the MetaMath-7B model.}
    \label{fig:MetaMath-7B-neurons-overlap}
\end{figure*}

\begin{figure*}[t!]
    \centering
    \input{Appendix-figures/MetaMath-13B-neurons-overlap}
    \caption{The overlap between non-English and English activated neurons and the normalized number of activated neurons across all languages in the MetaMath-13B model.}
    \label{fig:MetaMath-13B-neurons-overlap}
\end{figure*}

\begin{figure*}[t!]
    \centering
    \input{Appendix-figures/RFT-7B-neurons-overlap}
    \caption{The overlap between non-English and English activated neurons and the normalized number of activated neurons across all languages in the RFT-MuggleMath-7B model.}
    \label{fig:RFT-7B-neurons-overlap}
\end{figure*}

\begin{figure*}[t!]
    \centering
    \input{Appendix-figures/RFT-13B-neurons-overlap}
    \caption{The overlap between non-English and English activated neurons and the normalized number of activated neurons across all languages in the RFT-MuggleMath-13B model.}
    \label{fig:RFT-13B-neurons-overlap}
\end{figure*}

\end{document}